\documentclass[]{elsarticle}
\usepackage[a4paper, total={6in, 8in}, margin=0.9in]{geometry}
\usepackage{multicol}
\usepackage{amssymb}
\usepackage{graphicx}
\usepackage{booktabs}
\usepackage{multirow}
 
\usepackage{amsmath}
\usepackage{caption}
\usepackage{subcaption}
\usepackage{algorithm}
\usepackage{algpseudocode}

\usepackage{amsmath,amssymb,amsfonts}
\usepackage{algorithm}
\usepackage{algpseudocode}
\usepackage{graphicx}
\usepackage{textcomp}
\usepackage{multirow}
\usepackage{longtable}
\usepackage{rotating} 
\usepackage{longtable}
\usepackage{booktabs}
\usepackage{ltxtable} 
\usepackage{supertabular,booktabs}
\usepackage[export]{adjustbox}

\begin{document}

\begin{frontmatter}



\title{Data transformation based optimized customer churn prediction model for the telecommunication industry}


\author[ss]{Joydeb Kumar Sana}
 \address[ss]{Department of Computer Science and Engineering, Bangladesh University of Engineering and Technology, Dhaka, Bangladesh (e-mail: joysana@gmail.com)}
  

\author[bb]{Mohammad Zoynul Abedin}
  \address[bb]{Collaborative Innovation Center for Transport Studies, Dalian Maritime University, Dalian, China (e-mail: abedinmz@yahoo.com)}

\author[cc]{M. Sohel Rahman}
  \address[cc]{Department of Computer Science and Engineering, Bangladesh University of Engineering and Technology, Dhaka, Bangladesh (e-mail: sohel.kcl@gmail.com)}

\author[dd]{M. Saifur Rahman}
 \address[dd]{Department of Computer Science and Engineering, Bangladesh University of Engineering and Technology, Dhaka, Bangladesh (e-mail: saifur80@gmail.com)}

 \cortext[cor]{M. Saifur Rahman}
 \ead{saifur80@gmail.com}
 




\author{  }

\begin{abstract}
Data transformation (DT) is a process that transfers the original data into a form which supports a particular classification algorithm and helps to analyze the data for a special purpose. To improve the prediction performance we investigated various data transform methods. This study is conducted in a customer churn prediction (CCP) context in the telecommunication industry (TCI), where customer attrition is a common phenomenon. We have proposed a novel approach of combining data transformation methods with the machine learning models for the CCP problem. We conducted our experiments on publicly available TCI datasets and assessed the performance in terms of the widely used evaluation measures (e.g. AUC, precision, recall, and F-measure).  In this study, we presented comprehensive comparisons to affirm the effect of the transformation methods. The comparison results and statistical test  proved that most of the proposed data transformation based optimized models improve the performance of CCP significantly. Overall, an efficient and optimized CCP model for the telecommunication industry has been presented through this manuscript.
\end{abstract}


\end{frontmatter}


\section{Introduction}\label{sec:introduction}
Over the last few decades, the telecommunication industry (TCI) has witnessed enormous growth and development in terms of technology, level of competition, number of operators, new products and services and so on. However, because of extensive competition, saturated markets, dynamic environment, and attractive and lucrative offers, the TCI faces serious customer churn issues, which is considered to be a formidable problem in this regard \cite{b01}. In a competitive market, where customers have numerous choice of service providers, they can easily switch services and even service providers. Such customers are referred to as churned customers~\cite{b01} with respect to the original service provider. 

The three main generic strategies to generate more revenues in an industry are (i) to increase the retention period of customers, (ii) to acquire new customers and (iii) to up-sell the existing customers being the other two~\cite{b11}. In fact, customer retention is believed to be the most profitable strategy, as customer turnover severely hits the company’s income and its marketing expenses~\cite{b12_Amin_Anwar}.

Churn is an inevitable result of a customer’s long term dissatisfaction over the company’s services. Complete withdrawal from a service (provider) on part of a customer does not happen in a day; rather the dissatisfaction of the customer, grown over time and exacerbated by the lack of attention by the service provider, results in such a fiery gesture by the customer. To prevent this, the service provider must work on limitations (perceived by the customers) in its services to retain the aggrieved customers. Thus it is highly beneficial for a service provider to be able to identify a customer as a potential churned customer. 
In this context, non-churn customers are those who are reluctant to move from one service provider to another in contrast to churn customers.

If a telephone company (TELCO) can predict that a customer is likely to churn, then it can potentially cater targeted offerings to that customer to reduce his dissatisfaction, increase his engagement and thus potentially retain him/her. This has a clear positive impact on revenue. Additionally, customer churn adversely affects the company's fame and branding. 
As such, churn prediction is a very important task particularly in the telecom sector. To this end, TELCOs generally maintain a detailed standing report of the customer’s to understand their standing and to anticipate their longevity in continuing the services. Since the expense of getting new customers is relatively high~\cite{b7, b34}, TELCO nowadays principally focus on retaining their long-term customers rather than getting new ones. This makes churn prediction essential in the telecom sector~\cite{b8, b9}. With the above backdrop, in this paper, we revisit the customer churn prediction (CCP) problem as a binary classification problem in which all of the customers are partitioned into two classes, namely, Churn and Non-Churn.

\subsection{Brief Literature review}\label{sec:literature_review}
The problem of CCP has been tackled using various approaches including machine learning models, data mining methods, and hybrid techniques. 
Several Machine Learning (ML) and data mining approaches (e.g.,  Rough set theory \cite{b12_Amin_Anwar, b13_Amin_Shehzad}, Naïve Bayes and Bayesian network \cite{b14_Kirui}, Decision tree \cite{b18, b19}, Logistic regression \cite{b19}, RotBoost\cite{Idris_Khan_b19}, Support Vector Machine (SVM) \cite{Renjith_b19}, Genetic algorithm based neural network \cite{b15_Pendharkar}, AdaBoost Ensemble learning technique \cite{b16}, etc.) have been proposed for churn prediction in the TCI using customer relationship management (CRM) data. Notably, CRM data is widely used in prediction and classification problems \cite{b17}. A detailed literature review considering all these works is beyond the scope of this paper; however, we briefly review some of the most relevant papers below.

Brandusoiu et al.~\cite{Gavril_Toderean_b20} presented a data mining based approach for prepaid customer churn prediction. To reduce data dimension, the authors applied Principal Component Analysis (PCA). Three machine learning classifiers were used here, namely, Neural Networks (NN), Support Vector Machine (SVM), and Bayes Networks (BN) to predict churn customers. He et al.~\cite{b21} proposed a model based on Neural Networks (NN) in order to tackle the CCP problem in a large Chinese TELCO that had about $5.23$ million customers. Idris et al.~\cite{b22} proposed a technique combining genetic programming with AdaBoost to model the churn problem in the TCI. Huang et al.~\cite{b23} studied the problem of CCP in the big data platform. The aim of the study was to show that big data significantly improves the performance of churn prediction using Random Forest classifier.

Makhtar et al.~\cite{b24_Makhtar} proposed a rough set theory based model for churn prediction in TELCO. Amin et al.~\cite{b25_Amin_imbalance} on the other hand focused on tackling the data imbalance issue in the context of CCP in TELCO and compared six unique sampling strategies for oversampling. Burez et al.~\cite{b26_Burez} also studied the issue of unbalanced datasets in churn prediction models and conducted a comparative study for different methods for tackling the data imbalance issue. 
Hybrid strategies have also been used for processing massive amount of customer information together with regression techniques that provide effective churn prediction results~\cite{b22_Qureshi_Rehman}. On the other hand, Etaiwi et al.~\cite{b27_Etaiwi} showed that their Naïve Bayes model was able to beat SVM in terms of precision, recall, and F-measure. 

To the best of our knowledge, an important limitation in this context is that most of the methods in the literature have been experimented on a single dataset. Also, the impact of data transformation methods on CCP models have not been investigated deeply. 
There are various DT methods like the Log, Rank, Z-score, Discretization, Min-max, Box-cox, Aarcsine and so on. Among these, researchers broadly used the Log, Z-score, and Rank DT methods in different domains (e.g., software metrics normality and maintainability \cite{b27} \cite{b28}, defect prediction \cite{b29}, dimensionality reduction \cite{b29} etc.). To the best of our knowledge There are only one work in the literature where DT methods have been applied in the context of CCP in TELCO ~\cite{b38_Adnan_Babar}, where only two DT methods (e.g., Log and Rank) and a single classifier (e.g., Naïve Bayes) have been leveraged. Therefore, a large room for improvement is there in this context, which we consider in this work. 


\subsection{Our Contributions}
This paper makes the following key contributions:
\begin{itemize}
    \item We develop customer churn prediction models that leverage various data transformation (DT) methods and various optimized machine learning algorithms. In particular, we have combined six different DT methods with eight different optimized classifiers to develop a number of models to handle the CCP problem. The DT methods we utilized are: Log, Rank, Box-cox, Z-score, Discretization and Weight-of-evidence (WOE). On the other hand the classification algorithms we used include K-Nearest Neighbor (KNN), Naïve Bayes (NB), Logistic Regression (LR), Random forest (RF), Decision tree (DTree), Gradient boosting (GB), Feed-Forward Neural Networks (FNN) and Recurrent Neural Networks (RNN). 
    \item We have conducted extensive experiments on three different publicly available datasets and evaluated our models using various information retrieval metrics such as, AUC, Precision, Recall and F-measure. Our models achieved promising results and conclusively found that the DT methods have positive impact on CCP models.
    \item We also conduct statistical tests to check whether our findings are statistically significant or not. Our results clearly indicate that the impact of DT methods on the classifiers is not only positive but also statistically significant. 
\end{itemize}

\section{Materials and Methods}\label{sec:methodology}

\subsection{Datasets} \label{sec:Datasets} 
We use three publicly available benchmark datasets (referred to as Dataset- 1, 2 and 3 henceforth), that are broadly used for the CCP problem in the telecommunication area. Table \ref{table:dataset} describes these three datasets.

\begin{table}[h!]
\caption{Summary of datasets}
\label{table:dataset}
\begin{center}
\begin{tabular}{ p{4cm} p{2cm}  p{2cm}  p{2cm}  }
 \hline
 \vspace{.05mm}\\
 Description& Dataset-1 &Dataset-2&Dataset-3\\
 \vspace{.05mm}\\
 \hline
 \vspace{.01mm}\\
 No. of samples   & 100000    &5000&   3333\\
 No. of attributes&   101  & 20   &21\\
 No. of class labels & 2 & 2&  2\\
 Percentage churn samples    &50.43 &85.86& 85.5\\
 Percentage non-churn samples&   49.56  & 14.14&14.5\\
 Source of the datasets& URL$^1$  & URL$^2$   &URL$^3$\\
 \hline
\end{tabular}
\end{center}
URL$^1$: https://www.kaggle.com/abhinav89/telecom-customer/data (Last Access: September 29, 2019).\\
URL$^2$: https://data.world/earino/churn (Last Access: February 10, 2020).\\
URL$^3$: https://www.kaggle.com/becksddf/churn-in-telecoms-dataset/data (Last Access: February 10, 2020).

\end{table}

\subsubsection{Data preprocessing}\label{subsec:datapreparation} 
We apply the following essential data preprocessing steps:

\begin{itemize}

 \item We ignore the sample IDs and/or descriptive texts which are used only for informational purposes.

 \item Redundant attributes are removed.
 
 \item Missing numerical values are replaced with zero (0) and missing categorical values are treated as a separate category.

 \item We normalize the categorical values (such as `yes' or `no', `true' or `false') into 0s and 1s where each value represents the corresponding category \cite{b13_Amin_Shehzad}. Label encoder is used to normalize the categorical attributes. 
 
\end{itemize}

\subsection{Data Transformation (DT) Methods} \label{sec:dtMethods}
data transformation refers to the application of a deterministic mathematical function to each point in a data set. Table \ref{table:DTMethods} provides a description of the DT methods leveraged in our research.

 \begin{longtable}[c]{  p{2cm}  p{4cm} p{7cm}  }
 \caption{List of data transformation methods.\label{table:DTMethods}}\\
 \hline
 \multicolumn{3}{ c }{Begin of Table}\\
 \hline
DT Method & Description &  Equation\\
 \hline
 \endfirsthead

 \hline
 \multicolumn{3}{c}{Continuation of Table \ref{table:DTMethods}}\\
 \hline
DT Method & Description &  Equation\\
 \hline
 \endhead

 \hline
 \endfoot

 \hline
 \multicolumn{3}{ c }{End of Table}\\
 \hline\hline
 \endlastfoot

 \hline
 \vspace{.05mm}\\

 Log  &   Each variable x is replaced with log(x), where the base of the log is left up to the analyst \cite{b28} \cite{b39} \cite{b40}. In this study, In case the feature value contains zero, a constant 1 is typically added, along with ln(x)  & 
 \begin{equation}  \label{eq:log_transformation}
     \text{Log-DT(x)} = 
              \begin{cases}
               ln(x+1) & \text{if x =0} \\
               ln(x) & \text{if x>0} \\
               \end{cases}
\end{equation} where x is the value of any feature variable of the original dataset.  \\
  \vspace{.01mm}\\
 Rank &  It is a statistically calculated rank value \cite{b28} \cite{b41}. In this research, we followed the study \cite{b28} to transform the initial values of every feature in a original dataset into ten (10) ranks, using each 10th \% (percentile) of the given feature’s values  
 & 
  \begin{equation}  \label{eq:rank_transformation}
  Rank (x) =
    \begin{cases}
      1 & \text{if  x $\epsilon$ $[0,Q_1]$}\\
      k & \text{if  x $\epsilon$ $[Q_{(k-1)}, Q_k],   k \epsilon \{2, 3.....9\}$}\\
      10 & \text{ if x $\epsilon$ $[Q_9+\infty]$}
    \end{cases}       
\end{equation} where $Q_k$ is the $k \times 10$ percentile of the corresponding metric and symbol $\infty$ is the infinity.  \\
  \vspace{.01mm}\\
Box-Cox & It is a lamba based power transformation method \cite{b28} \cite{b40}. This transformation method is a process to transform non-normal dependent feature values into a normal distribution. 
& 
 \begin{equation} \label{eq:BoxCox_transformation}
     \text{Box-Cox}(x, \lambda)=\bigg\{\frac{x^2 - 1}{\lambda} \hspace{3mm}  \lambda \neq 0
\end{equation}

Where  $\lambda$ is configurable to the analyst, and x is the given value of any feature of the initial dataset. The $\lambda$ value = -5 to +5. In this study, we used $\lambda$ = 0.5.
\\
 \vspace{.01mm}\\
Z-score  &  It indicates the distance of a data point from the mean in units of standard deviation \cite{b42}.  
&
 \begin{equation} \label{eq:zscore_transformation}
     \text{Z-Score} =\frac{x - \text{sample mean}}{\text{sample standard deviation}}
\end{equation}
where x is the given value of any feature of the original dataset.
 \\
 \vspace{.01mm}\\
 Discretization& It is a binning technique \cite{b43}. For continuous variables, four widely used discretization techniques are K-means, equal width, equal frequency, and decision tree based discretization. We used the equal width discretization technique which is a very simple method.  
 & 
 For any given continuous variable x, the following process is applied: 
Provided $x_{min}$ is the minimum of a selected feature and $x_{max}$ is the maximum, bin width $\Omega$ can be computed as
  \begin{equation} \label{eq:discretization_2}
      \Omega=\frac{x_{max} - x_{min}}{b}
\end{equation}
Hence, the discretization technique generates b bins with boundaries at $x_{min+i} \times \Omega$, where i=1,2,.....(b-1). b is a parameter chosen by the analyst.
  \\
  \vspace{.01mm}\\
 Weight-of-evidence (WOE)&  It is binning and logarithmic based transformation\cite{b44}. Most of the cases, the WOE solves the skewed problem in the data distribution. WOE is the natural logarithm (ln) of the distribution which is the distribution of the good events (1) divided by the distribution of the bad events (0).
 &   
 \begin{equation} \label{eq:Weightofevidence}
       WOE = ln\bigg(\frac{\text{Distribution of churn customers}}{\text{Distribution of non-churn customers}}\bigg)
\end{equation}
  
 \\ 

 \hline
 
 \end{longtable}

\subsection{Evaluation Measures} \label{subsec:evaluation_measure}
The confusion matrix is generally used to assess the overall performance of a predictive model. For the CCP problem, the individual components of confusion matrix is defined as follows: (i) True Positives (TP): correctly predicted churn customers (ii) True Negatives (TN): correctly predicted non-churn customers (iii) False Positives (FP): non-churn customers, miss-predicted as churned customers and (iv) False Negatives (FN): churn customers, miss-predicted as non-churn customers. We use the following popular evaluation measures for comparing the performance of the models.

\textbf{Precision :}  Mathematically precision can be expressed as:
   \begin{equation} \label{eq:precision}
       Precision = \frac{TP}{TP+FP}
  \end{equation}

\textbf{The probability of detection (POD)/ Recall:} POD or recall is a valid choice of evaluation metric when we want to capture as many true churn customers as possible. Mathematically POD can be expressed as:
  \begin{equation} \label{eq:pod}
      Recall/ POD = \frac{TP}{FN+TP}
  \end{equation}

\textbf{The probability of false alarm (POF):} The value of POF should be small as much as possible (in an ideal case, the value of POF is 0 ). Mathematically POF can be defined as:
  \begin{equation} \label{eq:pod}
       POF/ \text{False positive rate} = \frac{FP}{TN+FP}
  \end{equation}
  We use POF for measuring incorrect churn predictions.

\textbf{The area under the curve (AUC):} Both POF and POD are used to measure the AUC \cite{b28} \cite{Amin_Adnan_b54}. A higher AUC value indicates a higher performance of the model. Mathematically AUC can be expressed as:

   \begin{equation} \label{eq:auc}
       AUC = \frac{1+POD-POF}{2}
  \end{equation}

\textbf{ F-Measure:} The F-measure is the harmonic mean of the precision and recall. F-measure is needed when we want to seek a balance between precision and recall. A perfect model has an F-measure of 1. The Mathematical formula of F-measure is defined below. 

  \begin{equation} \label{eq:f-measure}
       \text{F-Measure} =\frac{(2*precision*recall)}{(precision+recall)}
  \end{equation}


\begin{table}[h!]
\caption{List of baseline classifiers.}
\label{table:BaseLineClassifiers}
\begin{center}
\begin{tabular}{ p{0.7cm} p{2cm}  p{2cm}  p{8cm}  }
 \hline
 \vspace{.05mm}\\
 Key& Classifer & Model type& Description\\
 \vspace{.05mm}\\
 \hline
 \vspace{.01mm}\\
 KNN   & K-Nearest Neighbor & Instance-based learning, lazy learning &  The KNN algorithm assumes that similar things exist in close proximity.\\
 NB   & Naïve Bayes &Gaussian &  NB is a family of probabilistic algorithms. It gives the conditional probability, based on the Bayes theorem \\
 LR & Logistic Regression & Statistical model&  Logistic regression is estimating the parameters of a logistic model (a form of binary regression).\\
 RF&    Random forest  & Trees   & RF is an ensemble tree-based learning algorithm\\
 DTree&    Decision tree   & Trees   & DTree builds classification or regression models in the form of tree structure\\
 GB&   Gradient boosting   & Trees   & GB is an ensemble tree-based boosting method\\
 FNN&   Feed-Forward Networks  & Deep learning   & FNN is a deep learning classifier where the input travels in one direction\\
 RNN&    Recurrent Neural Networks & Deep learning   & RNN is a deep learning classifier where the output from previous step are fed as input to the current step. \\

 \hline
\end{tabular}
\end{center}
\end{table}

\subsection{Optimized CCP models} \label{subsec:optimized_CCP_model}
The baseline classifiers used in our research are presented in Table \ref{table:BaseLineClassifiers}. To examine the effect of the DT methods, we apply them on the original datasets and subsequently, on the transformed data, we train our CCP models with multiple machine learning classifiers (KNN, NB, LR, RF, DTree, GB, FNN and RNN) listed in Table \ref{table:BaseLineClassifiers}.

\subsubsection{Validation method and steps} \label{subsec:validataion_method}

 


In all our experiments, the classifiers of the CCP models were trained and tested using 10-fold cross-validation on the three different datasets described in Table \ref{table:dataset}. Firstly, a RAW data based CCP model was constructed without leveraging any of the DT methods on any features of the original datasets. In this case, we did not apply any feature selection steps either. However, we used the best hyper-parameters for the classifiers. 

Subsequently, we applied a DT method on each attribute of the dataset and retrained our models based on this transformed dataset. We experimented with each of the DT methods mentioned in Table~\ref{table:BaseLineClassifiers}. For each DT based model, we also used a feature selection and optimization procedure, which is described in the following section.

\begin{figure}[!htb]
\begin{center}
\includegraphics[height=400px,width=250px]{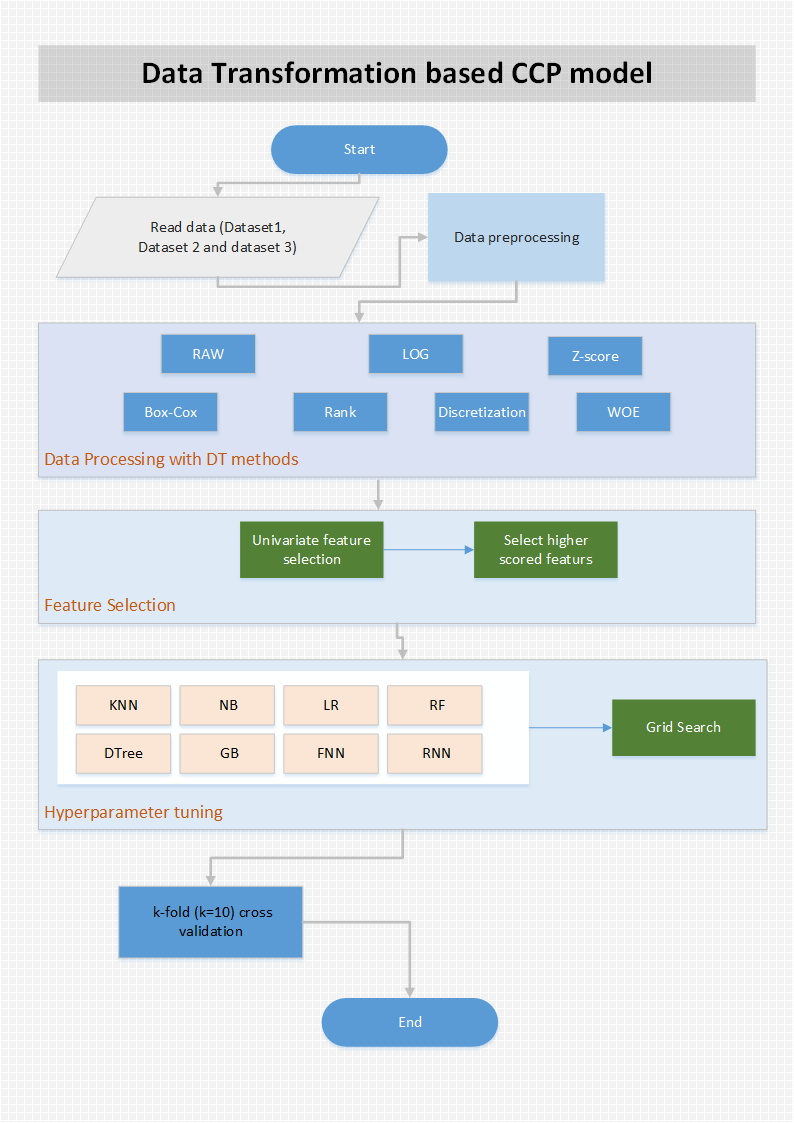}
\caption{Flowchart  of the Optimized CCP model using data transformation methods.}
\label{fig:DT_framework}
\end{center}
\end{figure}

\subsubsection{Feature Selection and Optimization} \label{subsec:feature_selection_optimization}
We have a set of hyper-parameters and we aim to find the right combination of the values thereof which will optimize the objective function. 
For tuning the hyper-parameters, we have applied grid search \cite{b52}. Figure \ref{fig:DT_framework} illustrates the overall flowchart of our proposed optimized CCP model. First, we applied some necessary preprocessing steps on the datasets. Then, DT methods (Log, Rank, Box-cox, Z-score, Discretization, and WOE ) were applied thereon. Next, we used the univariate feature selection technique to select the higher scored features from the dataset (we selected the top 80 features for dataset-1 and top 15 features for both dataset-2 and dataset-3). We applied grid search to find the best hyper-parameters for individual classifier algorithms. Finally, 10-fold cross validation was employed to train and validate the models.



\section{Stability measurement tests}  \label{sec:FriedmanTest}
We used Friedman non-parametric statistical test (FMT) \cite{b53} to examine the reliability of the findings and whether the improvement achieved by the DT based classification models are statistically significant. The Friedman test is the non-parametric statistical test for analyzing and finding differences in treatments across multiple attempts \cite{b53}. It does not assume any particular distribution of the data. Friedman test ranks all the methods. It ranks the classifiers independently for each dataset. Lower rank indicates a better performer.  
We performed the Friedman test on the F-measure results. Here, the null hypothesis $(H_0)$ represents: ``there is no difference among the performances of the CCP models". In our experiments, the test was carried out with the significance level, $\alpha = 0.05$.

Subsequently, post hoc Holm test is conducted to perform the paired comparisons with respect to the best performing DT model. In particular, when the null hypothesis is rejected, we used the post hoc Holm test to compare the performance of the models. This test is a similarity measurement process that compares all the models. We performed the Holm's post hoc comparison for $\alpha = 0.05$ and $\alpha = 0.10$.

\section{DT methods and Data Distribution}  \label{sec:result_analysis}
Data transformation attempts to change the data from one representation to another to enhance the quality thereof with a goal to enable analysis of certain information for specific purposes. In order to find out the impact of the DT methods on the datasets, data skewness and data normality measurement tests have been performed on the three different datasets and the results are visualized through Q-Q (quantile-quantile) plots \cite{Amin_Adnan_b54, b28}. 

\subsubsection{Coding and Experimental Environment}
All experiments were conducted on a machine having Windows 10, 64-bit system with Intel Core i7 3.6GHz processor, 24GB RAM, and 500GB HD. All codes were implemented with Python 3.7. Jupyter Notebook was used for coding. All data and code are available at the following link:  {https://github.com/joysana1/Churn-prediction}.



\section{Results} \label{sec:result_discussion} 



\begin{figure}[tb]
\begin{center}
\begin{subfigure}[t]{1.0\hsize}
\centering
 \includegraphics[height=150px,width=250px]{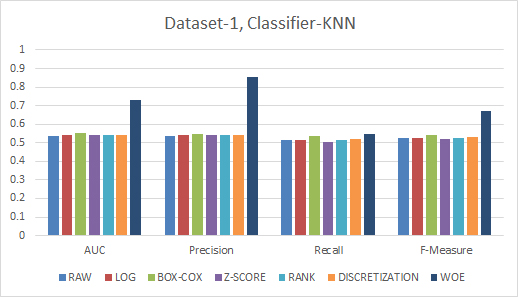}
\caption{Performance comparison, Dataset-1}
\label{fig:D-1-KNN}
\end{subfigure}   
\begin{subfigure}[t]{1.0\hsize}
\centering
  \includegraphics[height=150px,width=250px]{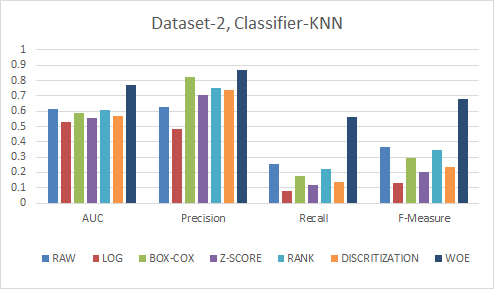}
\caption{Performance comparison, Dataset-2}
\label{fig:D-2-KNN}
\end{subfigure}

\begin{subfigure}[t]{1.0\hsize}
\centering
  \includegraphics[height=150px,width=250px]{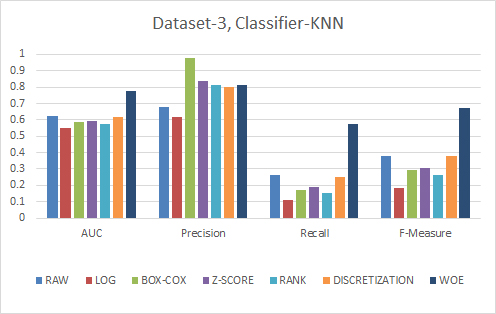}
\caption{Performance comparison, Dataset-3}
\label{fig:D-3-KNN}
\end{subfigure}

    \caption{Performance comparison among the CCP methods using KNN classifier}
  \label{fig:KNN_classifier}
\end{center}
\end{figure}
\begin{figure}[tb]
\begin{center}
\begin{subfigure}[t]{1.0\hsize}
\centering
 \includegraphics[height=150px,width=250px]{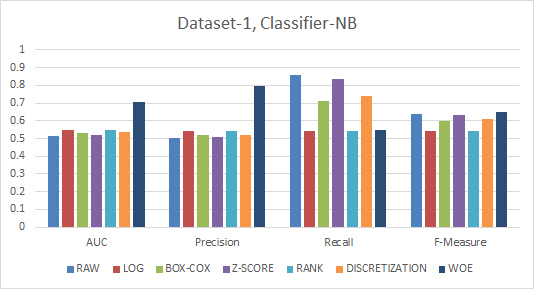}
\caption{Performance comparison, Dataset-1 }
\label{fig:D-1-NB}
\end{subfigure}   
\begin{subfigure}[t]{1.0\hsize}
\centering
 \includegraphics[height=150px,width=250px]{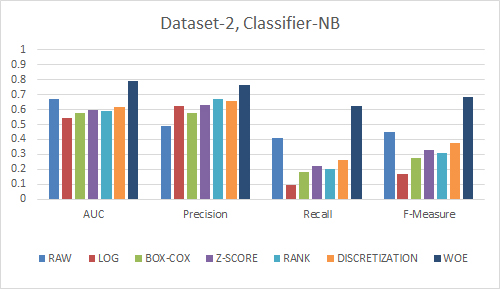}
\caption{Performance comparison, Dataset-2}
\label{fig:D-2-NB}
\end{subfigure}

\begin{subfigure}[t]{1.0\hsize}
\centering
 \includegraphics[height=150px,width=250px]{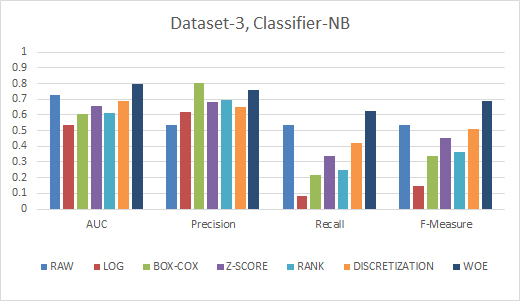}
\caption{Performance comparison, Dataset-3}
\label{fig:D-3-NB}
\end{subfigure}

    \caption{Performance comparison among the DT methods using NB classifier}
  \label{fig:NB_classifier}
\end{center}
\end{figure}
\begin{figure}[tb]
\begin{center}
\begin{subfigure}[t]{1.0\hsize}
\centering
\includegraphics[height=150px,width=250px]{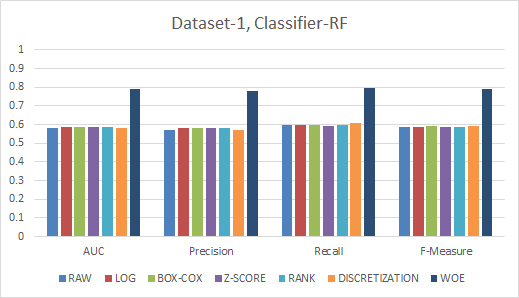}
\caption{Performance comparison, Dataset-1}
\label{fig:D-1-RF}
\end{subfigure}   
\begin{subfigure}[t]{1.0\hsize}
\centering
 \includegraphics[height=150px,width=250px]{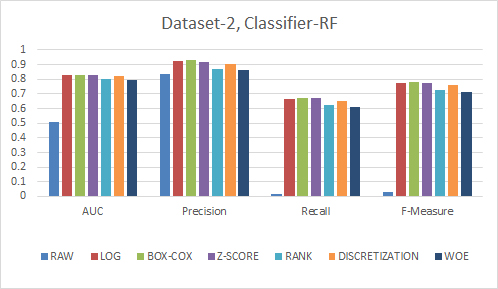}
\caption{Performance comparison, Dataset-2}
\label{fig:D-2-RF}
\end{subfigure}

\begin{subfigure}[t]{1.0\hsize}
\centering
\includegraphics[height=150px,width=250px]{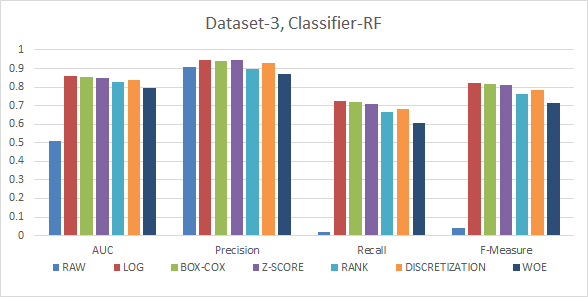}
\caption{Performance comparison, Dataset-3}
\label{fig:D-3-RF}
\end{subfigure}

    \caption{Performance comparison among the DT methods using RF classifier}
  \label{fig:RF_classifier}
\end{center}
\end{figure}
\begin{figure}[tb]
\begin{center}
\begin{subfigure}[t]{1.0\hsize}
\centering
\includegraphics[height=150px,width=250px]{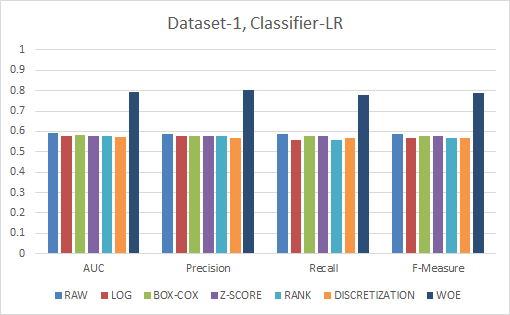}
\caption{Performance comparison, Dataset-1}
\label{fig:D-1-LR}
\end{subfigure}   
\begin{subfigure}[t]{1.0\hsize}
\centering
 \includegraphics[height=150px,width=250px]{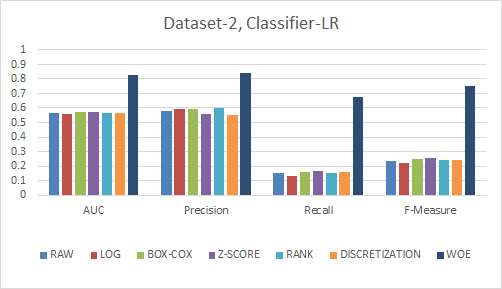}
\caption{Performance comparison, Dataset-2}
\label{fig:D-2-LR}
\end{subfigure}

\begin{subfigure}[t]{1.0\hsize}
\centering
\includegraphics[height=150px,width=250px]{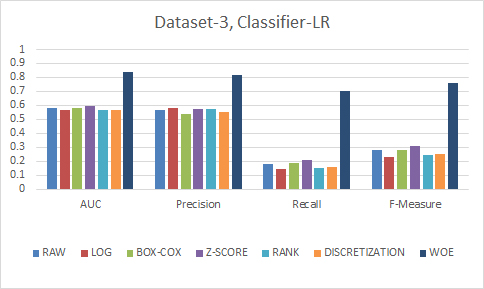}
\caption{Performance comparison, Dataset-3}
\label{fig:D-3-LR}
\end{subfigure}

    \caption{Performance comparison among the DT methods using LR classifier}
  \label{fig:LR_classifier}
\end{center}
\end{figure}
\begin{figure}[tb]
\begin{center}
\begin{subfigure}[t]{1.0\hsize}
\centering
\includegraphics[height=150px,width=250px]{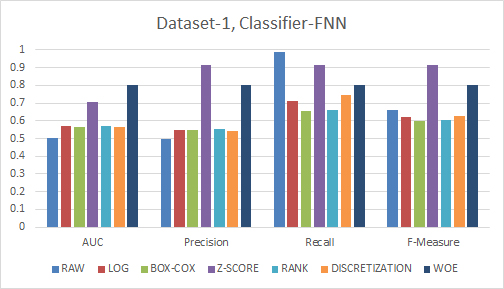}
\caption{Performance comparison, Dataset-1}
\label{fig:D-1-FNN}
\end{subfigure}   
\begin{subfigure}[t]{1.0\hsize}
\centering
 \includegraphics[height=150px,width=250px]{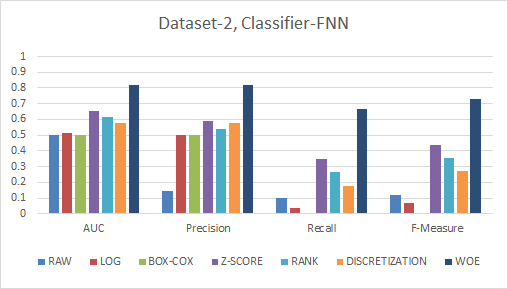}
\caption{Performance comparison, Dataset-2}
\label{fig:D-2-FNN}
\end{subfigure}

\begin{subfigure}[t]{1.0\hsize}
\centering
\includegraphics[height=150px,width=250px]{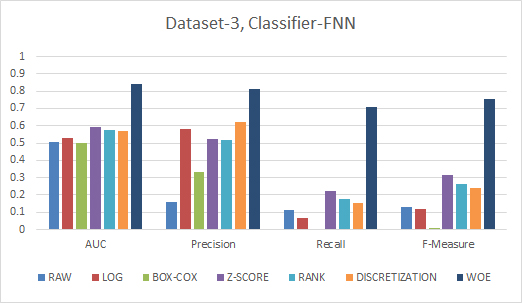}
\caption{Performance comparison, Dataset-3}
\label{fig:D-3-FNN}
\end{subfigure}

    \caption{Performance comparison among the DT methods using FNN classifier}
  \label{fig:FNN_classifier}
\end{center}
\end{figure}
\begin{figure}[tb]
\begin{center}
\begin{subfigure}[t]{1.0\hsize}
\centering
\includegraphics[height=150px,width=250px]{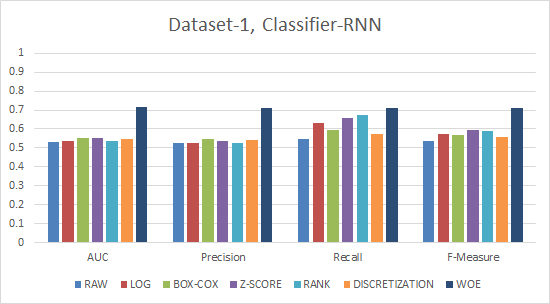}
\caption{Performance comparison, Dataset-1}
\label{fig:D-1-RNN}
\end{subfigure}   
\begin{subfigure}[t]{1.0\hsize}
\centering
\includegraphics[height=150px,width=250px]{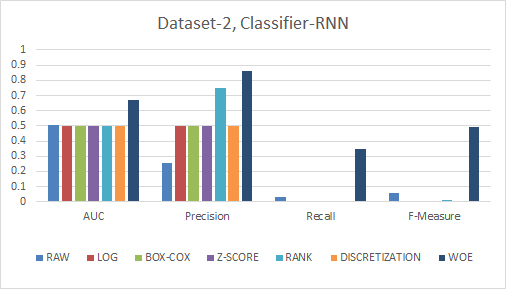}
\caption{Performance comparison, Dataset-2}
\label{fig:D-2-RNN}
\end{subfigure}

\begin{subfigure}[t]{1.0\hsize}
\centering
\includegraphics[height=150px,width=250px]{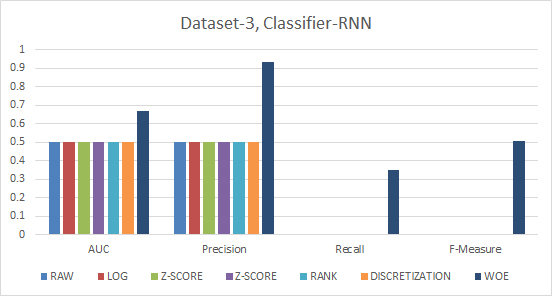}
\caption{Performance comparison, Dataset-3}
\label{fig:D-3-RNN}
\end{subfigure}

    \caption{Performance comparison among the DT methods using RNN classifier}
  \label{fig:RNN_classifier}
\end{center}
\end{figure}
\begin{figure}[tb]
\begin{center}
\begin{subfigure}[t]{1.0\hsize}
\centering
\includegraphics[height=150px,width=250px]{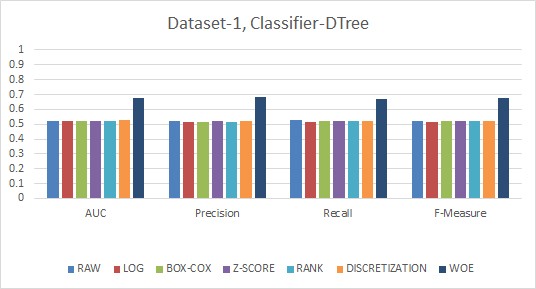}
\caption{performance comparison, Dataset-1}
\label{fig:D-1-DT}
\end{subfigure}   
\begin{subfigure}[t]{1.0\hsize}
\centering
\includegraphics[height=150px,width=250px]{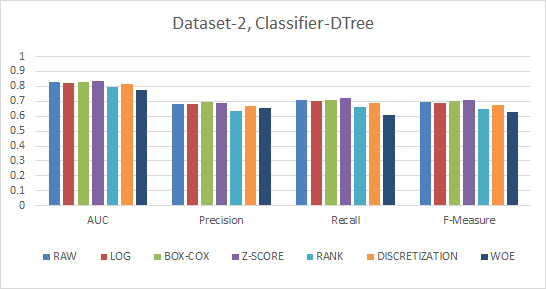}
\caption{performance comparison, Dataset-2}
\label{fig:D-2-DT}
\end{subfigure}

\begin{subfigure}[t]{1.0\hsize}
\centering
\includegraphics[height=150px,width=250px]{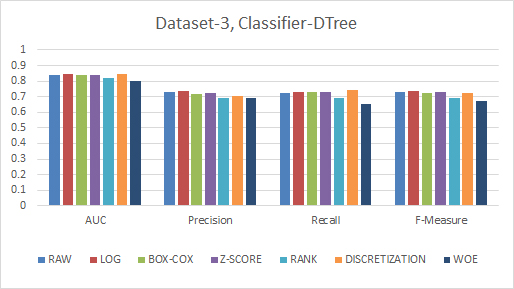}
\caption{performance comparison, Dataset-3}
\label{fig:D-3-DT}
\end{subfigure}

    \caption{performance comparison among the DT methods using DTree classifier}
  \label{fig:DTree_classifier}
\end{center}
\end{figure}
\begin{figure}[tb]
\begin{center}
\begin{subfigure}[t]{1.0\hsize}
\centering
\includegraphics[height=150px,width=250px]{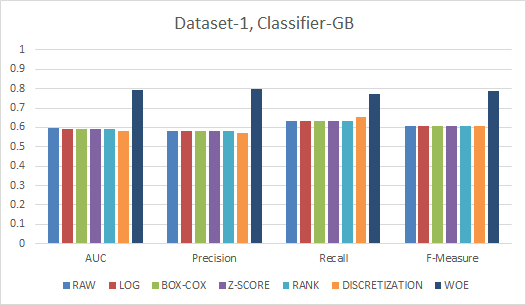}
\caption{performance comparison, Dataset-1}
\label{fig:D-1-GB}
\end{subfigure}   
 
\begin{subfigure}[t]{1.0\hsize}
\centering
\includegraphics[height=150px,width=250px]{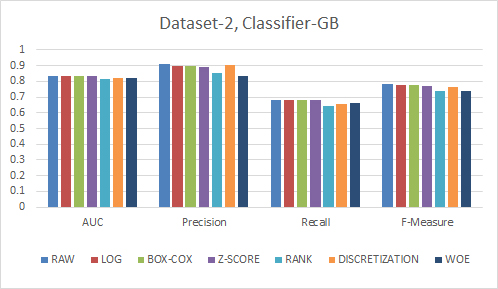}
\caption{performance comparison, Dataset-2}
\label{fig:D-2-GB}
\end{subfigure}
 
\begin{subfigure}[t]{1.0\hsize}
\centering
\includegraphics[height=150px,width=250px]{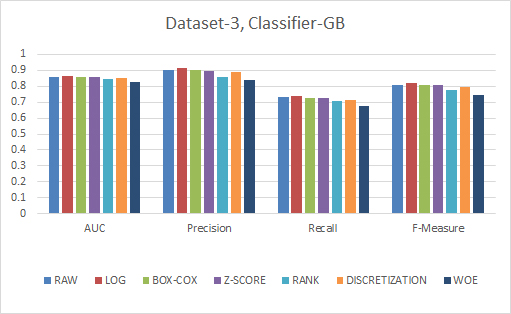}
\caption{performance comparison, Dataset-3}
\label{fig:D-3-GB}
\end{subfigure}

    \caption{performance comparison among the DT methods using GB classifier}
  \label{fig:GB_classifier}
\end{center}
\end{figure}

The impact of the DT methods on all the 8 classifiers (through rigorous experimentation on 3 benchmark datasets) are illustrated in Figures \ref{fig:KNN_classifier} through \ref{fig:GB_classifier}. Each of these figures illustrates the performance comparison (in terms of AUC, precision, recall, and F-measure) among the RAW data based CCP model and other DT methods based CCP models for all three datasets as follows (please check Table \ref{table:map} for a map for understanding the figures). Table \ref{table:dt_methods_on_dataset-1}, \ref{table:dt_methods_on_dataset-2} and \ref{table:dt_methods_on_dataset-3} in the supplementary file reports the values for all the measures for all the datasets.

 \twocolumn

\onecolumn


\begin{table}[htbp!]
\centering
\renewcommand{\arraystretch}{1.5}
\begin{tabular}{|l|l|l|l|}
\hline
 \textbf{Figure} & \textbf{Classifier} & \textbf{Sub-figure} & \textbf{Dataset}   \\ \hline
\multirow{3}{*}{1} & \multirow{3}{*}{KNN} & a  & 1       \\ \cline{3-4} 
                   &                      & b  & 2       \\ \cline{3-4} 
                   &                      & c  & 3       \\ \hline
\multirow{3}{*}{2} & \multirow{3}{*}{NB}  & a  & 1       \\ \cline{3-4} 
                   &                      & b  & 2       \\ \cline{3-4} 
                   &                      & c  & 3       \\ \hline
\multirow{3}{*}{3} & \multirow{3}{*}{RF}  & a  & 1       \\ \cline{3-4} 
                   &                      & b  & 2       \\ \cline{3-4} 
                   &                      & c  & 3       \\ \hline          
\multirow{3}{*}{4} & \multirow{3}{*}{LR}  & a  & 1       \\ \cline{3-4} 
                   &                      & b  & 2       \\ \cline{3-4} 
                   &                      & c  & 3       \\ \hline              
\multirow{3}{*}{5} & \multirow{3}{*}{FNN}  & a  & 1       \\ \cline{3-4} 
                   &                      & b  & 2       \\ \cline{3-4} 
                   &                      & c  & 3       \\ \hline              
\multirow{3}{*}{6} & \multirow{3}{*}{RNN}  & a  & 1       \\ \cline{3-4} 
                   &                      & b  & 2       \\ \cline{3-4} 
                   &                      & c  & 3       \\ \hline              
\multirow{3}{*}{7} & \multirow{3}{*}{DTree}  & a  & 1       \\ \cline{3-4} 
                   &                      & b  & 2       \\ \cline{3-4} 
                   &                      & c  & 3       \\ \hline              
\multirow{3}{*}{8} & \multirow{3}{*}{GB}  & a  & 1       \\ \cline{3-4} 
                   &                      & b  & 2       \\ \cline{3-4} 
                   &                      & c  & 3       \\ \hline              
\end{tabular}
 \caption{Map of the results illustrated in different figures}
    \label{table:map}
\end{table}

\subsection{Results on Dataset 1}
The performance of the baseline classifiers (referred to as RAW in the figures) in dataset 1 is quite poor in all the metrics: the best performer in terms of F-measure is NB with a value of 0.636 only. Interestingly, not all DT methods performed better than RAW. However, the performance of WOE is consistently better than RAW across all classifiers. In a few cases of course some other DT methods able to outperform WOE: for example, across all combinations in Dataset 1, the best individula performance is achieved by FNN with Z-SCORE with a staggering F-Measure of 0.917. As for AUC as well, the most consistent performer is WOE with the best value achieved for FNN (0.802)

\subsection{Results on Dataset 2}
Interestingly, the performance of some baseline classifiers in Dataset 2 is quite impressive in Dataset 2, particularly in the context of AUC. For example, both DT and GB (RAW version) achieved more than 0.82 as AUC; the F-Measure was also acceptable, particularly for GB (0.78). 

Among the DT methods, again, WOE performs (in terms of F-Measure) most consistently albeit with the glitch that for DT and GB, it performs slightly worse than RAW. In fact, surprisingly enough, for GB, the best performer is RAW; for DT however, Z-SCORE is the winner, very closely followed by BOX-COX.        

\subsection{Results on Dataset 3}
In Dataset 3 as well, the performance of DT and GB in RAW mode is quite impressive: for DT the AUC and F-Measure values are respectively 0.84 and 0.727 and for GB these are even better, 0.86 and 0.809, respectively. Again, the performance of WOE is the most consistent except in the case of DT and GB where it is beaten by RAW. The overall winner is GB with LOG transformation which registers 0.864 as AUC and 0.818 as F-Measure.

\section{Statistical test results}  \label{sec:FriedmanTest}
\begin{table}[h!]
 \centering
 \caption{Average Rankings of the algorithms}
\label{table:FMT_RANK}
{
\renewcommand{\arraystretch}{1.5}
\begin{tabular}{p{3cm} p{2.5cm} }\hline
\textbf{Algorithm } & \textbf{Rank (\#Position) }\\
\hline
 WOE & 2.4167 (\#1)\\
 Z-SCORE & 3.5417 (\#2)\\
RAW & 3.7917 (\#3)\\
Discritization & 4.0833 (\#4)\\
BOX-COX & 4.1667 (\#5)\\
RANK & 4.9375  (\#6)\\
LOG & 5.0625  (\#7)\\
\hline
\end{tabular}
}
\end{table}

\begin{figure}[!htb]
\begin{center}
\includegraphics[height=200px,width=250px]{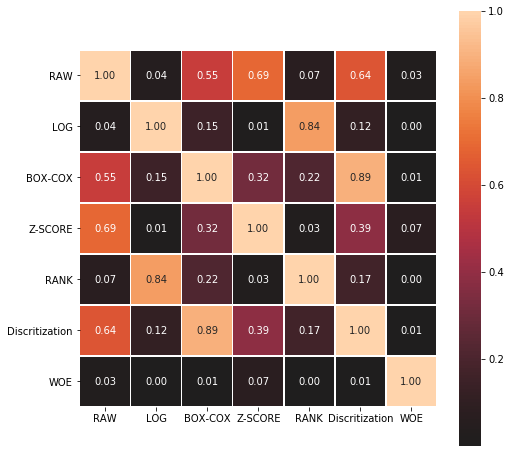}
\caption{Performance difference heatmap among DT based CCP models in terms of p-value}
\label{fig:heatmap-Average-F-measure-P-value}
\end{center}
\end{figure}

Table \ref{table:FMT_RANK} summarizes the ranking of the Freedman test among the DT methods. Friedman statistic distributed according to Chi-square with ($n$-1) degrees of freedom is 24.700893. Here $n$ is the number of methods. P-value computed by Friedman test is 0.00039. Form the Chi-square distribution table, critical value is 12.59. Notably, $99.5\%$ confidence interval (CI) has been considered for this test. Our Friedman test statistic value (24.700893) is greater than the critical value (12.59). So the decision is to reject the null hypothesis $(H_0)$. Subsequently, the post hoc Holm test revealed significant differences among the DT methods. Figure \ref{fig:heatmap-Average-F-measure-P-value} illustrates the results of Holm's test as a heat map. \textit{p}-value $\le0.05$ was considered as the evidence of significance. Figure \ref{fig:heatmap-Average-F-measure-P-value} tells that WOE performance is significantly different from other DT methods except for the Z-SCORE. Table \ref{table:FMT_posthoc_1} reflects the post hoc comparisons for $\alpha = 0.05$ and $\alpha = 0.10$. When the p-value of the test is smaller than the significant rate $\alpha$ = 10\% and 5\% then Holm's procedure rejects the null hypothesis. Evidently, WOE DT based models are found to be significantly better than the other models.

\begin{table}[hbt!] 
\begin{center}
\caption{ Friedman and Holm test result }
\label{table:FMT_posthoc_1}
{
\renewcommand{\arraystretch}{1.5}
\begin{tabular}{ p{0.1cm} p{3cm} p{2cm}  p{2cm}  p{2cm} }
\hline
$i$&Method& p-value &Hypothesis ($\alpha=0.05$) & Hypothesis ($\alpha = 0.10$)\\ 
\hline1&WOE vs. LOG  &0.000022&Rejected&Rejected\\
2&WOE vs. RANK&0.000053&Rejected&Rejected\\
3&WOE vs. BOX-COX&0.005012&Rejected&Rejected\\
4&WOE vs. Discritization&0.007526&Rejected&Rejected\\
5&WOE vs. RAW&0.027461&Rejected&Rejected\\
6&WOE vs. Z-SCORE&0.071229&Not Rejected&Rejected\\
\hline
 \vspace{.01mm}\\
\end{tabular}
}
\end{center}
\end{table}


 

\section{Impact of the DT methods on Data Distribution}  \label{sec:result_analysis}
The Q-Q plots are shown in Figure \ref{fig:q_q_plot_dataset_1}, \ref{fig:q_q_plot_dataset_2} and \ref{fig:q_q_plot_dataset_3} for Dataset-1, Dataset-2 and Dataset-3, respectively. As we found WOE and Z-Score DT methods are performing better than the RAW (without DT) method (see the Friedman ranked table \ref{table:FMT_RANK}), we generated Q-Q plots only for RAW, WOE, and Z-Score methods. In each Q-Q plot, the first 3 features of the respective dataset are shown. From the Q-Q plots, it is observed that after transformation by the WOE DT method, we achieved less skewness (i.e., the data became more normally distributed). Normally distributed data is beneficial for  the classifiers~\cite{Amin_Adnan_b54, COUSSEMENT201727}. Similar performance is also observed for Z-SCORE. 

\section{Discussion}  \label{sec:discussion}
From the comparative analysis and statistical tests, it is evident that DT methods have a great impact on improving the CCP performance in TELCO. A few prior works (e.g., \cite{b27}, \cite{b28}, and \cite{b38_Adnan_Babar}) also studied the effect of DT methods but in a limited scale and did not consider the optimization issues. We on the other hand conducted a comprehensive study considering six DT methods and eight machine learning classifiers on three different benchmark datasets. The performance of the DT based classifiers have been investigated in terms of AUC, precision, recall, and F-measure.

\begin{figure}[tb]
\begin{center}
\begin{subfigure}[t]{1.0\hsize}
\centering
\includegraphics[height=140px,width=400px, frame]{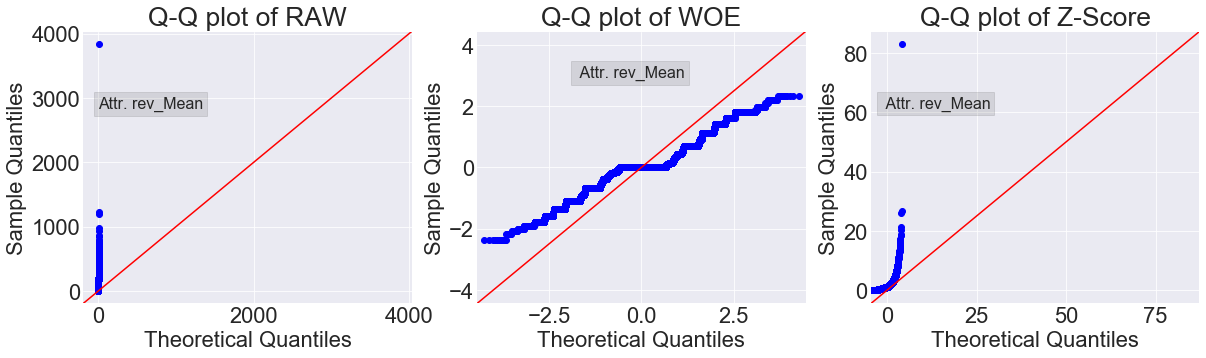}
\includegraphics[height=140px,width=400px, frame]{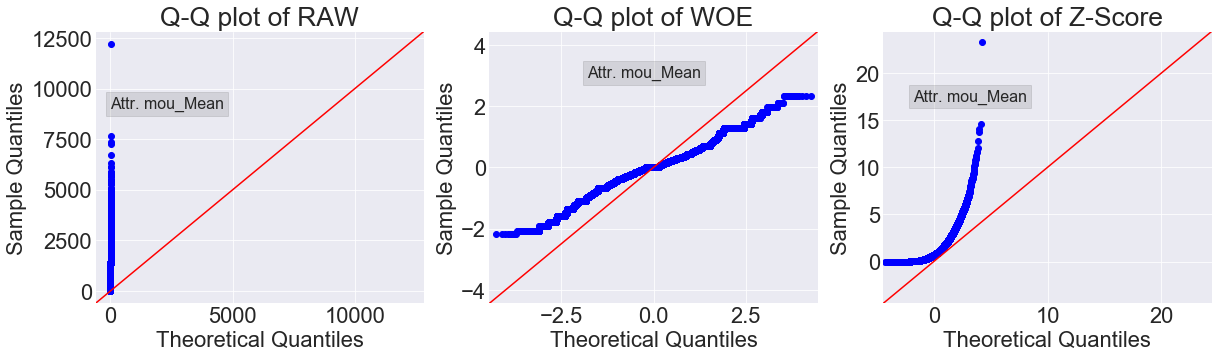}
\includegraphics[height=140px,width=400px, frame]{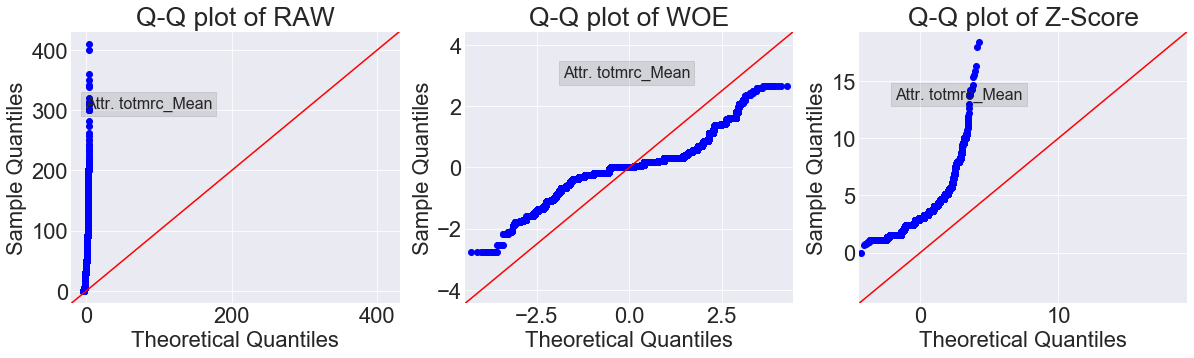}
\label{fig:q-q_plot_woe}
\end{subfigure}   
\caption{The Q-Q plot for WOE, Z-Score DT method and without DT method on dataset-1 }
\label{fig:q_q_plot_dataset_1}
\end{center}
\end{figure}

\begin{figure}[tb]
\begin{center}

\begin{subfigure}[t]{1.0\hsize}
\centering
\includegraphics[height=140px,width=400px, frame]{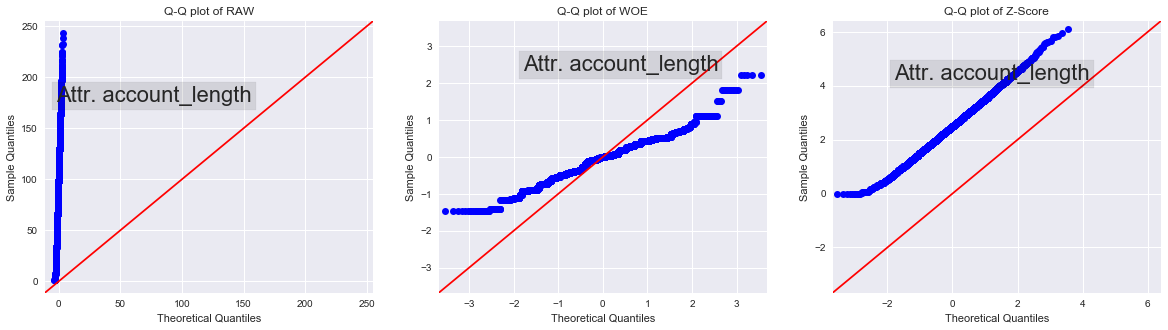}
\includegraphics[height=140px,width=400px, frame]{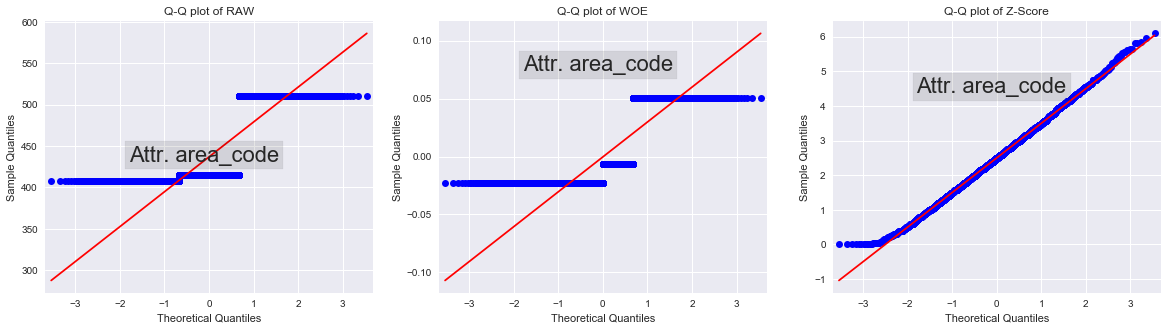}
\includegraphics[height=140px,width=400px, frame]{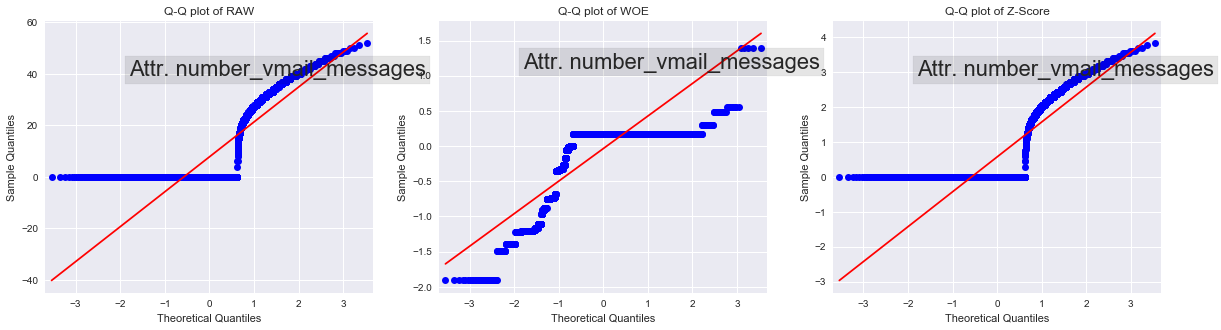}
\end{subfigure}   
\caption{The Q-Q plot for WOE, Z-Score DT method and without DT method on dataset-2 }
\label{fig:q_q_plot_dataset_2}
\end{center}
\end{figure}

\begin{figure}[tb]
\begin{center}

\begin{subfigure}[t]{1.0\hsize}
\centering
 \includegraphics[height=140px,width=400px, frame]{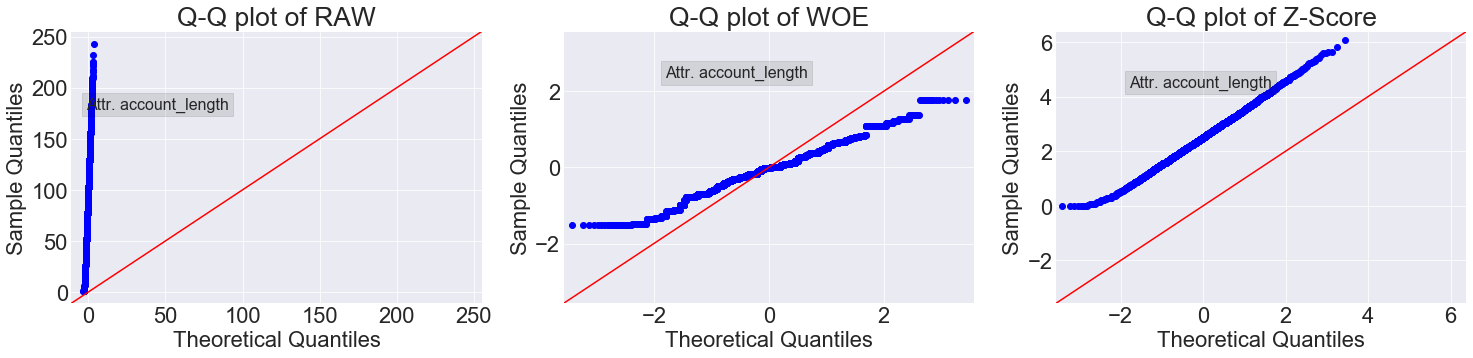}
 \includegraphics[height=140px,width=400px, frame]{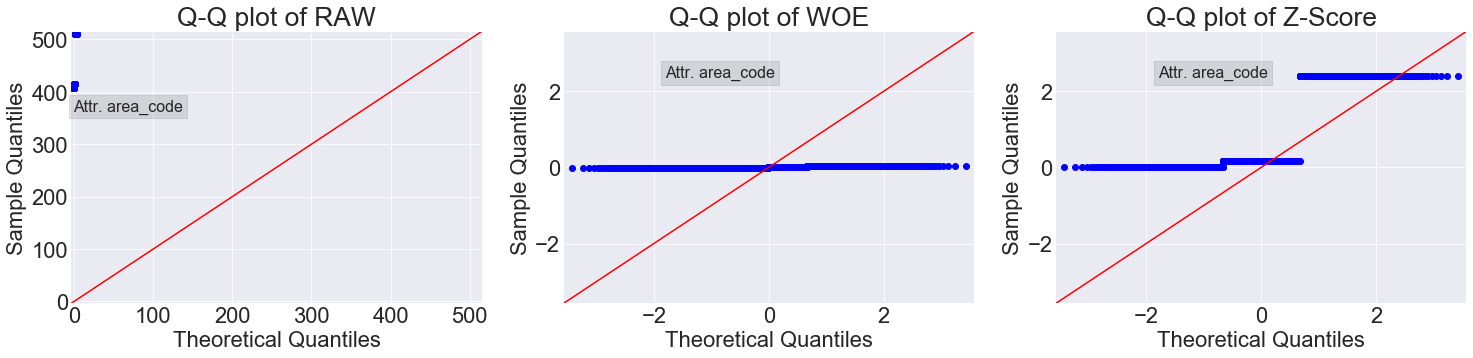}
\includegraphics[height=140px,width=400px, frame]{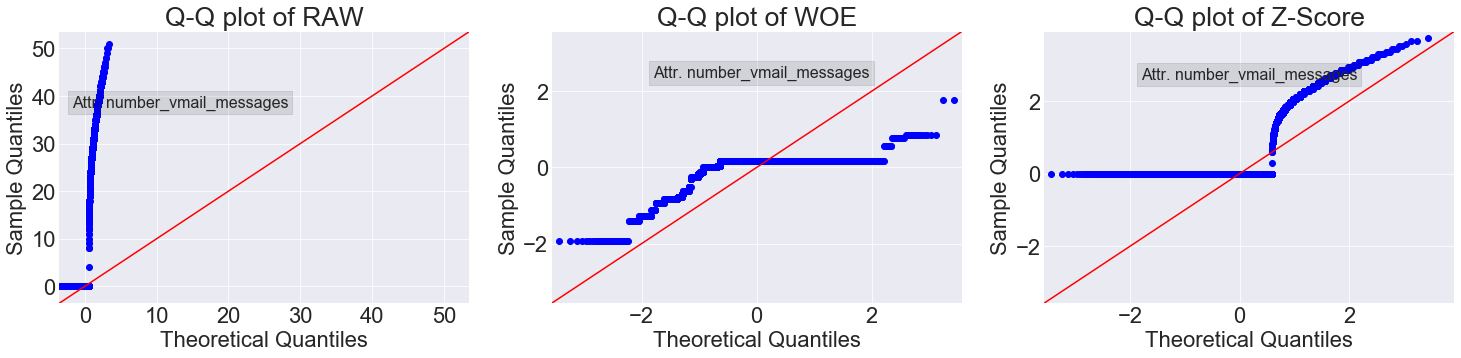}
\end{subfigure}   
\caption{The Q-Q plot for WOE, Z-Score DT method and without DT method on dataset-3 }
\label{fig:q_q_plot_dataset_3}
\end{center}
\end{figure}

The data transformation techniques have shown great promise in improving the data distribution quality in general. Specially, in our experiments, the WOE method improved the data normality which in the sequel provided a clear positive impact on the prediction performance for the customer churn prediction (Figures \ref{fig:q_q_plot_dataset_1} - \ref{fig:q_q_plot_dataset_3}).

The comparative analyses involving the RAW based and DT based CCP models clearly suggested the potential of DT methods in improving the CCP performance (Figures \ref{fig:KNN_classifier} through \ref{fig:GB_classifier}). In particular, our experimental results strongly suggested that the WOE method contributed a lot towards improving the performance, albeit with the exception of DTree and GB classifiers for the datasets 2 and 3. While the performance of WOE in these cases satisfactory, it failed to outperform RAW based model performance. We hypothesize that this is due to the binning technique within the WOE method. Moreover, those two datasets are unbalanced datasets. The DTree and GB classifiers might consider them as some order which is not a specific order.

From Table \ref{table:FMT_RANK} we notice that WOE is the best ranked method and the rank value is 2.4167. The post hoc comparison heatmap \ref{fig:heatmap-Average-F-measure-P-value} and Table \ref{table:FMT_posthoc_1} reflect how the WOE is better than the other methods. As Friedman test is rejecting the null hypothesis $(H_0)$ and post hoc Holm analysis advocates the WOE method's supremacy, it is clear that  DT methods improve the user churn prediction performance significantly for the telecommunication industry. Therefore, to construct a successful CCP model, we recommend to select the best classifier (LR, FNN) and the WOE data transfer method.
 
 
\section{Conclusion}  \label{sec:conclusion}
Predicting customer churn is one of the most important factors in business planning in TELCOs. To improve the churn prediction performance we investigated with six different data transformation methods, namely, Log, Rank, Box-cox, Z-score, Discretization, and Weight-of-evidence. We used eight different machine learning classifiers which are K-Nearest neighbor (KNN), Naïve Bayes (NB), Logistic regression (LR), Random forest (RF), Decision tree (DTree), Gradient boosting (GB), Feed-forward neural networks (FNN), Recurrent neural networks (RNN). For each classifier, we applied univariate feature selection method to select top ranked features and used grid search for hyper-parameter tuning. We evaluated our methods in terms of AUC, precision, recall, and F-measure. The experimental outcomes indicate that, in most cases, the data transformation methods enhance the data quality and improve the prediction performance. To support our experimental results we performed Friedman non-parametric statistical test and post hoc Holm statistical analysis. The Friedman statistical test and post hoc Holm statistical analysis confirmed that Weight-of-evidence and Z-score DT based CCP models perform better than the raw based CCP model. To test the robustness of our DT-augmented CCP models, we performed our experiments on both balanced (dataset-1) and non-balanced datasets (dataset-2 and dataset-3). CCP is still a hard and swiftly developing problem usually for competitive businesses and particularly for telecommunication companies. Future research is probably capable to offer higher outcomes on other datasets with multiple classifiers. Another future direction can be to extend this study with other types of data transformation approaches and classifiers. Our proposed model can be tested on the other telecom datasets to examine the generalization of our results at a larger scale. Last but not the least, work can be done to extend our approach to customer churn datasets from other business sectors to study the generalization of our claim across business domains.

\onecolumn

\clearpage

\bibliographystyle{elsarticle-harv}
\bibliography{Bibliography}

\begin{thebibliography}{39}
\expandafter\ifx\csname natexlab\endcsname\relax\def\natexlab#1{#1}\fi
\providecommand{\url}[1]{\texttt{#1}}
\providecommand{\href}[2]{#2}
\providecommand{\path}[1]{#1}
\providecommand{\DOIprefix}{doi:}
\providecommand{\ArXivprefix}{arXiv:}
\providecommand{\URLprefix}{URL: }
\providecommand{\Pubmedprefix}{pmid:}
\providecommand{\doi}[1]{\href{http://dx.doi.org/#1}{\path{#1}}}
\providecommand{\Pubmed}[1]{\href{pmid:#1}{\path{#1}}}
\providecommand{\bibinfo}[2]{#2}
\ifx\xfnm\relax \def\xfnm[#1]{\unskip,\space#1}\fi
\bibitem[{Amin et~al.(2016a)Amin, Anwar, Adnan, Nawaz, Aloufi, Hussain and
  Huang}]{b12_Amin_Anwar}
\bibinfo{author}{Amin, A.}, \bibinfo{author}{Anwar, S.},
  \bibinfo{author}{Adnan, A.}, \bibinfo{author}{Nawaz, M.},
  \bibinfo{author}{Aloufi, K.}, \bibinfo{author}{Hussain, A.},
  \bibinfo{author}{Huang, K.}, \bibinfo{year}{2016}a.
\newblock \bibinfo{title}{Customer churn prediction in telecommunication sector
  using rough set approach}.
\newblock \bibinfo{journal}{Neurocomputing}
  \DOIprefix\doi{10.1016/j.neucom.2016.12.009}.
\bibitem[{Amin et~al.(2016b)Amin, Anwar, Adnan, Nawaz, Howard, Qadir, Hawalah
  and Hussain}]{b25_Amin_imbalance}
\bibinfo{author}{Amin, A.}, \bibinfo{author}{Anwar, S.},
  \bibinfo{author}{Adnan, A.}, \bibinfo{author}{Nawaz, M.},
  \bibinfo{author}{Howard, N.}, \bibinfo{author}{Qadir, J.},
  \bibinfo{author}{Hawalah, A.}, \bibinfo{author}{Hussain, A.},
  \bibinfo{year}{2016}b.
\newblock \bibinfo{title}{Comparing oversampling techniques to handle the class
  imbalance problem: A customer churn prediction case study}.
\newblock \bibinfo{journal}{IEEEAccess} \bibinfo{volume}{PP},
  \bibinfo{pages}{7940--7957}.
\newblock \DOIprefix\doi{10.1109/ACCESS.2016.2619719}.
\bibitem[{{Amin} et~al.(2018){Amin}, {Shah}, {Khattak}, {Baker}, u.~{Rahman
  Durani} and {Anwar}}]{b38_Adnan_Babar}
\bibinfo{author}{{Amin}, A.}, \bibinfo{author}{{Shah}, B.},
  \bibinfo{author}{{Khattak}, A.M.}, \bibinfo{author}{{Baker}, T.},
  \bibinfo{author}{u.~{Rahman Durani}, H.}, \bibinfo{author}{{Anwar}, S.},
  \bibinfo{year}{2018}.
\newblock \bibinfo{title}{Just-in-time customer churn prediction: With and
  without data transformation}, in: \bibinfo{booktitle}{2018 IEEE Congress on
  Evolutionary Computation (CEC)}, pp. \bibinfo{pages}{1--6}.
\bibitem[{Amin et~al.(2019)Amin, Shah, Khattak, {Lopes Moreira}, Ali, Rocha and
  Anwar}]{Amin_Adnan_b54}
\bibinfo{author}{Amin, A.}, \bibinfo{author}{Shah, B.},
  \bibinfo{author}{Khattak, A.M.}, \bibinfo{author}{{Lopes Moreira}, F.J.},
  \bibinfo{author}{Ali, G.}, \bibinfo{author}{Rocha, A.},
  \bibinfo{author}{Anwar, S.}, \bibinfo{year}{2019}.
\newblock \bibinfo{title}{Cross-company customer churn prediction in
  telecommunication: A comparison of data transformation methods}.
\newblock \bibinfo{journal}{International Journal of Information Management}
  \bibinfo{volume}{46}, \bibinfo{pages}{304 -- 319}.
\newblock \URLprefix
  \url{http://www.sciencedirect.com/science/article/pii/S0268401218305930},
  \DOIprefix\doi{https://doi.org/10.1016/j.ijinfomgt.2018.08.015}.
\bibitem[{Amin et~al.(2015)Amin, Shehzad, Khan and Anwar}]{b13_Amin_Shehzad}
\bibinfo{author}{Amin, A.}, \bibinfo{author}{Shehzad, S.},
  \bibinfo{author}{Khan, C.}, \bibinfo{author}{Anwar, S.},
  \bibinfo{year}{2015}.
\newblock \bibinfo{title}{Churn prediction in telecommunication industry using
  rough set approach}, pp. \bibinfo{pages}{83--95}.
\newblock \DOIprefix\doi{10.1007/978-3-319-10774-5-8}.
\bibitem[{Bishara and Hittner(2015)}]{b41}
\bibinfo{author}{Bishara, A.J.}, \bibinfo{author}{Hittner, J.B.},
  \bibinfo{year}{2015}.
\newblock \bibinfo{title}{Reducing bias and error in the correlation
  coefficient due to nonnormality}.
\newblock \bibinfo{journal}{Educational and Psychological Measurement}
  \bibinfo{volume}{75}, \bibinfo{pages}{785--804}.
\newblock \DOIprefix\doi{10.1177/0013164414557639}.
\bibitem[{Brandusoiu et~al.(2016)Brandusoiu, Toderean and
  Beleiu}]{Gavril_Toderean_b20}
\bibinfo{author}{Brandusoiu, I.}, \bibinfo{author}{Toderean, G.},
  \bibinfo{author}{Beleiu, H.}, \bibinfo{year}{2016}.
\newblock \bibinfo{title}{Methods for churn prediction in the pre-paid mobile
  telecommunications industry}, pp. \bibinfo{pages}{97--100}.
\newblock \DOIprefix\doi{10.1109/ICComm.2016.7528311}.
\bibitem[{Burez and {Van den Poel}(2009)}]{b26_Burez}
\bibinfo{author}{Burez, J.}, \bibinfo{author}{{Van den Poel}, D.},
  \bibinfo{year}{2009}.
\newblock \bibinfo{title}{{Handling class imbalance in customer churn
  prediction}}.
\newblock \bibinfo{journal}{Expert Systems with Applications}
  \bibinfo{volume}{36}, \bibinfo{pages}{4626--4636}.
\newblock \URLprefix \url{http://dx.doi.org/10.1016/j.eswa.2008.05.027},
  \DOIprefix\doi{10.1016/j.eswa.2008.05.027}.
\bibitem[{Cheadle et~al.(2003)Cheadle, Vawter, Freed and Becker}]{b42}
\bibinfo{author}{Cheadle, C.}, \bibinfo{author}{Vawter, M.},
  \bibinfo{author}{Freed, W.}, \bibinfo{author}{Becker, K.},
  \bibinfo{year}{2003}.
\newblock \bibinfo{title}{Analysis of microarray data using z-score
  transformation}.
\newblock \bibinfo{journal}{The Journal of molecular diagnostics : JMD}
  \bibinfo{volume}{5}, \bibinfo{pages}{73--81}.
\newblock \DOIprefix\doi{10.1016/S1525-1578(10)60455-2}.
\bibitem[{Coussement et~al.(2017)Coussement, Lessmann and
  Verstraeten}]{COUSSEMENT201727}
\bibinfo{author}{Coussement, K.}, \bibinfo{author}{Lessmann, S.},
  \bibinfo{author}{Verstraeten, G.}, \bibinfo{year}{2017}.
\newblock \bibinfo{title}{A comparative analysis of data preparation algorithms
  for customer churn prediction: A case study in the telecommunication
  industry}.
\newblock \bibinfo{journal}{Decision Support Systems} \bibinfo{volume}{95},
  \bibinfo{pages}{27 -- 36}.
\newblock \URLprefix
  \url{http://www.sciencedirect.com/science/article/pii/S0167923616302020},
  \DOIprefix\doi{https://doi.org/10.1016/j.dss.2016.11.007}.
\bibitem[{De~Caigny et~al.(2018)De~Caigny, Coussement and De~Bock}]{b19}
\bibinfo{author}{De~Caigny, A.}, \bibinfo{author}{Coussement, K.},
  \bibinfo{author}{De~Bock, K.}, \bibinfo{year}{2018}.
\newblock \bibinfo{title}{A new hybrid classification algorithm for customer
  churn prediction based on logistic regression and decision trees}.
\newblock \bibinfo{journal}{European Journal of Operational Research}
  \bibinfo{volume}{269}.
\newblock \DOIprefix\doi{10.1016/j.ejor.2018.02.009}.
\bibitem[{Dem{\v{s}}ar(2006)}]{b53}
\bibinfo{author}{Dem{\v{s}}ar, J.}, \bibinfo{year}{2006}.
\newblock \bibinfo{title}{{Statistical comparisons of classifiers over multiple
  data sets}}.
\newblock \bibinfo{journal}{Journal of Machine Learning Research}
  \bibinfo{volume}{7}, \bibinfo{pages}{1--30}.
\bibitem[{Etaiwi et~al.(2017)Etaiwi, Biltawi and Naymat}]{b27_Etaiwi}
\bibinfo{author}{Etaiwi, W.}, \bibinfo{author}{Biltawi, M.},
  \bibinfo{author}{Naymat, G.}, \bibinfo{year}{2017}.
\newblock \bibinfo{title}{Evaluation of classification algorithms for banking
  customer’s behavior under apache spark data processing system}.
\newblock \bibinfo{journal}{Procedia Computer Science} \bibinfo{volume}{113},
  \bibinfo{pages}{559 -- 564}.
\newblock \URLprefix
  \url{http://www.sciencedirect.com/science/article/pii/S1877050917316897},
  \DOIprefix\doi{https://doi.org/10.1016/j.procs.2017.08.280}.
\bibitem[{Fayyad and Irani(1992)}]{b43}
\bibinfo{author}{Fayyad, U.M.}, \bibinfo{author}{Irani, K.B.},
  \bibinfo{year}{1992}.
\newblock \bibinfo{title}{On the handling of continuous-valued attributes in
  decision tree generation}.
\newblock \bibinfo{journal}{Machine Learning} \bibinfo{volume}{8},
  \bibinfo{pages}{87--102}.
\bibitem[{Feng et~al.(2014)Feng, Hongyue, Lu, Chen, He, Lu and Tu}]{b40}
\bibinfo{author}{Feng, C.}, \bibinfo{author}{Hongyue, W.}, \bibinfo{author}{Lu,
  N.}, \bibinfo{author}{Chen, T.}, \bibinfo{author}{He, H.},
  \bibinfo{author}{Lu, Y.}, \bibinfo{author}{Tu, X.}, \bibinfo{year}{2014}.
\newblock \bibinfo{title}{Log-transformation and its implications for data
  analysis}.
\newblock \bibinfo{journal}{Shanghai archives of psychiatry}
  \bibinfo{volume}{26}, \bibinfo{pages}{105--9}.
\newblock \DOIprefix\doi{10.3969/j.issn.1002-0829.2014.02.009}.
\bibitem[{Fukushima et~al.(2014)Fukushima, Kamei, McIntosh, Yamashita and
  Ubayashi}]{b29}
\bibinfo{author}{Fukushima, T.}, \bibinfo{author}{Kamei, Y.},
  \bibinfo{author}{McIntosh, S.}, \bibinfo{author}{Yamashita, K.},
  \bibinfo{author}{Ubayashi, N.}, \bibinfo{year}{2014}.
\newblock \bibinfo{title}{An empirical study of just-in-time defect prediction
  using cross-project models}.
\newblock \bibinfo{journal}{Empirical Software Engineering}
  \bibinfo{volume}{21}, \bibinfo{pages}{172--181}.
\newblock \DOIprefix\doi{10.1145/2597073.2597075}.
\bibitem[{Hadden et~al.(2008)Hadden, Tiwari, Roy and Ruta}]{b34}
\bibinfo{author}{Hadden, J.}, \bibinfo{author}{Tiwari, A.},
  \bibinfo{author}{Roy, R.}, \bibinfo{author}{Ruta, D.}, \bibinfo{year}{2008}.
\newblock \bibinfo{title}{{Churn prediction: Does technology matter}}.
\newblock \bibinfo{journal}{World Academy of Science, Engineering and
  Technology} , \bibinfo{pages}{973--979}\URLprefix
  \url{https://waset.org/journals/waset/v16/v16-162.pdf}.
\bibitem[{He et~al.(2009)He, He and Zhang}]{b21}
\bibinfo{author}{He, Y.}, \bibinfo{author}{He, Z.}, \bibinfo{author}{Zhang,
  D.}, \bibinfo{year}{2009}.
\newblock \bibinfo{title}{A study on prediction of customer churn in fixed
  communication network based on data mining}, pp. \bibinfo{pages}{92--94}.
\newblock \DOIprefix\doi{10.1109/FSKD.2009.767}.
\bibitem[{Huang et~al.(2010)Huang, Kechadi, Buckley, Kiernan, Keogh and
  Rashid}]{b17}
\bibinfo{author}{Huang, B.}, \bibinfo{author}{Kechadi, T.M.},
  \bibinfo{author}{Buckley, B.}, \bibinfo{author}{Kiernan, G.},
  \bibinfo{author}{Keogh, E.}, \bibinfo{author}{Rashid, T.},
  \bibinfo{year}{2010}.
\newblock \bibinfo{title}{A new feature set with new window techniques for
  customer churn prediction in land-line telecommunications}.
\newblock \bibinfo{journal}{Expert Systems with Applications}
  \bibinfo{volume}{37}, \bibinfo{pages}{3657 -- 3665}.
\newblock \URLprefix
  \url{http://www.sciencedirect.com/science/article/pii/S0957417409008914},
  \DOIprefix\doi{https://doi.org/10.1016/j.eswa.2009.10.025}.
\bibitem[{Huang et~al.(2015)Huang, Zhu, Yuan, Deng, Li, Ni, Dai, Yang and
  Zeng}]{b23}
\bibinfo{author}{Huang, Y.}, \bibinfo{author}{Zhu, F.}, \bibinfo{author}{Yuan,
  M.}, \bibinfo{author}{Deng, K.}, \bibinfo{author}{Li, Y.},
  \bibinfo{author}{Ni, B.}, \bibinfo{author}{Dai, W.}, \bibinfo{author}{Yang,
  Q.}, \bibinfo{author}{Zeng, J.}, \bibinfo{year}{2015}.
\newblock \bibinfo{title}{Telco churn prediction with big data}, pp.
  \bibinfo{pages}{607--618}.
\newblock \DOIprefix\doi{10.1145/2723372.2742794}.
\bibitem[{Hung et~al.(2006)Hung, Yen and Wang}]{b18}
\bibinfo{author}{Hung, S.Y.}, \bibinfo{author}{Yen, D.}, \bibinfo{author}{Wang,
  H.Y.}, \bibinfo{year}{2006}.
\newblock \bibinfo{title}{Applying data mining to telecom chum management}.
\newblock \bibinfo{journal}{Expert Systems with Applications}
  \bibinfo{volume}{31}, \bibinfo{pages}{515--524}.
\newblock \DOIprefix\doi{10.1016/j.eswa.2005.09.080}.
\bibitem[{Idris et~al.(2017)Idris, Iftikhar and Rehman}]{b16}
\bibinfo{author}{Idris, A.}, \bibinfo{author}{Iftikhar, A.},
  \bibinfo{author}{Rehman, Z.}, \bibinfo{year}{2017}.
\newblock \bibinfo{title}{Intelligent churn prediction for telecom using
  gp-adaboost learning and pso undersampling}.
\newblock \bibinfo{journal}{Cluster Computing} \bibinfo{volume}{22},
  \bibinfo{pages}{7241--7255}.
\bibitem[{Idris and Khan(2012)}]{Idris_Khan_b19}
\bibinfo{author}{Idris, A.}, \bibinfo{author}{Khan, A.}, \bibinfo{year}{2012}.
\newblock \bibinfo{title}{Customer churn prediction for telecommunication:
  Employing various various features selection techniques and tree based
  ensemble classifiers}, pp. \bibinfo{pages}{23--27}.
\newblock \DOIprefix\doi{10.1109/INMIC.2012.6511498}.
\bibitem[{Idris et~al.(2012)Idris, Khan and Lee}]{b22}
\bibinfo{author}{Idris, A.}, \bibinfo{author}{Khan, A.}, \bibinfo{author}{Lee,
  Y.S.}, \bibinfo{year}{2012}.
\newblock \bibinfo{title}{Genetic programming and adaboosting based churn
  prediction for telecom}, pp. \bibinfo{pages}{1328--1332}.
\newblock \DOIprefix\doi{10.1109/ICSMC.2012.6377917}.
\bibitem[{Keramati et~al.(2014)Keramati, Jafari-Marandi, Aliannejadi, Ahmadian,
  Mozaffari and Abbasi}]{b8}
\bibinfo{author}{Keramati, A.}, \bibinfo{author}{Jafari-Marandi, R.},
  \bibinfo{author}{Aliannejadi, M.}, \bibinfo{author}{Ahmadian, I.},
  \bibinfo{author}{Mozaffari, M.}, \bibinfo{author}{Abbasi, U.},
  \bibinfo{year}{2014}.
\newblock \bibinfo{title}{Improved churn prediction in telecommunication
  industry using data mining techniques}.
\newblock \bibinfo{journal}{Applied Soft Computing} \bibinfo{volume}{24},
  \bibinfo{pages}{994 -- 1012}.
\newblock \URLprefix
  \url{http://www.sciencedirect.com/science/article/pii/S1568494614004062},
  \DOIprefix\doi{https://doi.org/10.1016/j.asoc.2014.08.041}.
\bibitem[{Kirui et~al.(2013)Kirui, Hong, Cheruiyot and Kirui}]{b14_Kirui}
\bibinfo{author}{Kirui, C.}, \bibinfo{author}{Hong, L.},
  \bibinfo{author}{Cheruiyot, W.}, \bibinfo{author}{Kirui, H.},
  \bibinfo{year}{2013}.
\newblock \bibinfo{title}{Predicting customer churn in mobile telephony
  industry using probabilistic classifiers in data mining}.
\newblock \bibinfo{journal}{IJCSI Int. J. Comput. Sci. Issues}
  \bibinfo{volume}{10}, \bibinfo{pages}{165--172}.
\bibitem[{Lu and Ph(2002)}]{b7}
\bibinfo{author}{Lu, J.}, \bibinfo{author}{Ph, D.}, \bibinfo{year}{2002}.
\newblock \bibinfo{title}{{Predicting Customer Churn in the Telecommunications
  Industry –– An Application of Survival Analysis Modeling Using SAS}}.
\newblock \bibinfo{journal}{Techniques} \bibinfo{volume}{114-27},
  \bibinfo{pages}{114--27}.
\newblock \URLprefix \url{http://www2.sas.com/proceedings/sugi27/p114-27.pdf}.
\bibitem[{Makhtar et~al.(2017)Makhtar, Nafis, Mohamed, Awang, Rahman and
  Mat~Deris}]{b24_Makhtar}
\bibinfo{author}{Makhtar, M.}, \bibinfo{author}{Nafis, s.},
  \bibinfo{author}{Mohamed, M.A.}, \bibinfo{author}{Awang, M.K.},
  \bibinfo{author}{Rahman, M.}, \bibinfo{author}{Mat~Deris, M.},
  \bibinfo{year}{2017}.
\newblock \bibinfo{title}{Churn classification model for local
  telecommunication company based on rough set theory}.
\newblock \bibinfo{journal}{Journal of Fundamental and Applied Sciences}
  \bibinfo{volume}{9}, \bibinfo{pages}{854--68}.
\newblock \DOIprefix\doi{10.4314/jfas.v9i6s.64}.
\bibitem[{{Menzies} et~al.(2007){Menzies}, {Dekhtyar}, {Distefano} and
  {Greenwald}}]{b39}
\bibinfo{author}{{Menzies}, T.}, \bibinfo{author}{{Dekhtyar}, A.},
  \bibinfo{author}{{Distefano}, J.}, \bibinfo{author}{{Greenwald}, J.},
  \bibinfo{year}{2007}.
\newblock \bibinfo{title}{Problems with precision: A response to "comments on
  'data mining static code attributes to learn defect predictors'"}.
\newblock \bibinfo{journal}{IEEE Transactions on Software Engineering}
  \bibinfo{volume}{33}, \bibinfo{pages}{637--640}.
\bibitem[{Pendharkar(2009)}]{b15_Pendharkar}
\bibinfo{author}{Pendharkar, P.C.}, \bibinfo{year}{2009}.
\newblock \bibinfo{title}{Genetic algorithm based neural network approaches for
  predicting churn in cellular wireless network services}.
\newblock \bibinfo{journal}{Expert Systems with Applications}
  \bibinfo{volume}{36}, \bibinfo{pages}{6714 -- 6720}.
\newblock \URLprefix
  \url{http://www.sciencedirect.com/science/article/pii/S0957417408005757},
  \DOIprefix\doi{https://doi.org/10.1016/j.eswa.2008.08.050}.
\bibitem[{Renjith(2017)}]{Renjith_b19}
\bibinfo{author}{Renjith, S.}, \bibinfo{year}{2017}.
\newblock \bibinfo{title}{B2c e-commerce customer churn management: Churn
  detection using support vector machine and personalized retention using
  hybrid recommendations}.
\newblock \bibinfo{journal}{International Journal on Future Revolution in
  Computer Science and Communication Engineering (IJFRCSCE)}
  \bibinfo{volume}{3}, \bibinfo{pages}{34 – 39}.
\newblock \DOIprefix\doi{10.6084/M9.FIGSHARE.5579482}.
\bibitem[{S.~A.~Qureshi et~al.(2013)S.~A.~Qureshi, Rehman, Qamar, Kamal and
  Rehman}]{b22_Qureshi_Rehman}
\bibinfo{author}{S.~A.~Qureshi, S.}, \bibinfo{author}{Rehman, A.},
  \bibinfo{author}{Qamar, A.}, \bibinfo{author}{Kamal, A.},
  \bibinfo{author}{Rehman, A.}, \bibinfo{year}{2013}.
\newblock \bibinfo{title}{Telecommunication subscribers' churn prediction model
  using machine learning}, pp. \bibinfo{pages}{131--136}.
\newblock \DOIprefix\doi{10.1109/ICDIM.2013.6693977}.
\bibitem[{Siddiqi(2005)}]{b44}
\bibinfo{author}{Siddiqi, N.}, \bibinfo{year}{2005}.
\newblock \bibinfo{title}{Credit risk scorecards: Developing and implementing
  intelligent credit scoring}, \bibinfo{publisher}{Wiley}.
\bibitem[{Syarif et~al.(2016)Syarif, Prugel-Bennett and Wills}]{b52}
\bibinfo{author}{Syarif, I.}, \bibinfo{author}{Prugel-Bennett, A.},
  \bibinfo{author}{Wills, G.}, \bibinfo{year}{2016}.
\newblock \bibinfo{title}{Svm parameter optimization using grid search and
  genetic algorithm to improve classification performance}.
\newblock \bibinfo{journal}{TELKOMNIKA (Telecommunication Computing Electronics
  and Control)} \bibinfo{volume}{14}, \bibinfo{pages}{1502}.
\newblock \DOIprefix\doi{10.12928/telkomnika.v14i4.3956}.
\bibitem[{Wei and Chiu(2002)}]{b11}
\bibinfo{author}{Wei, C.P.}, \bibinfo{author}{Chiu, I.T.},
  \bibinfo{year}{2002}.
\newblock \bibinfo{title}{Turning telecommunications call details to churn
  prediction: A data mining approach}.
\newblock \bibinfo{journal}{Expert Systems with Applications}
  \bibinfo{volume}{23}, \bibinfo{pages}{103--112}.
\newblock \DOIprefix\doi{10.1016/S0957-4174(02)00030-1}.
\bibitem[{Xie et~al.(2009)Xie, Li, Ngai and Ying}]{b9}
\bibinfo{author}{Xie, Y.}, \bibinfo{author}{Li, X.}, \bibinfo{author}{Ngai,
  E.}, \bibinfo{author}{Ying, W.}, \bibinfo{year}{2009}.
\newblock \bibinfo{title}{Customer churn prediction using improved balanced
  random forests}.
\newblock \bibinfo{journal}{Expert Systems with Applications}
  \bibinfo{volume}{36}, \bibinfo{pages}{5445 -- 5449}.
\newblock \URLprefix
  \url{http://www.sciencedirect.com/science/article/pii/S0957417408004326},
  \DOIprefix\doi{https://doi.org/10.1016/j.eswa.2008.06.121}.
\bibitem[{Zhang et~al.(2017)Zhang, Keivanloo and Zou}]{b28}
\bibinfo{author}{Zhang, F.}, \bibinfo{author}{Keivanloo, I.},
  \bibinfo{author}{Zou, Y.}, \bibinfo{year}{2017}.
\newblock \bibinfo{title}{Data transformation in cross-project defect
  prediction}.
\newblock \bibinfo{journal}{Empirical Software Engineering}
  \bibinfo{volume}{22}, \bibinfo{pages}{1--33}.
\newblock \DOIprefix\doi{10.1007/s10664-017-9516-2}.
\bibitem[{Zhang et~al.(2013)Zhang, Mockus, Zou, Khomh and Hassan}]{b27}
\bibinfo{author}{Zhang, F.}, \bibinfo{author}{Mockus, A.},
  \bibinfo{author}{Zou, Y.}, \bibinfo{author}{Khomh, F.},
  \bibinfo{author}{Hassan, A.E.}, \bibinfo{year}{2013}.
\newblock \bibinfo{title}{How does context affect the distribution of software
  maintainability metrics?}, pp. \bibinfo{pages}{350--359}.
\newblock \DOIprefix\doi{10.1109/ICSM.2013.46}.
\bibitem[{Óskarsdóttir et~al.(2017)Óskarsdóttir, Bravo, Verbeke, Sarraute,
  Baesens and Vanathien}]{b01}
\bibinfo{author}{Óskarsdóttir, M.}, \bibinfo{author}{Bravo, C.},
  \bibinfo{author}{Verbeke, W.}, \bibinfo{author}{Sarraute, C.},
  \bibinfo{author}{Baesens, B.}, \bibinfo{author}{Vanathien, J.},
  \bibinfo{year}{2017}.
\newblock \bibinfo{title}{Social network analytics for churn prediction in
  telco: Model building, evaluation and network architecture}.
\newblock \bibinfo{journal}{Expert Systems with Applications}
  \bibinfo{volume}{85}.
\newblock \DOIprefix\doi{10.1016/j.eswa.2017.05.028}.

\end{thebibliography}

 \appendix
\begin{table}[]
\begin{center}
\caption{RAW and  DT method based CCP models on dataset-1}
\label{table:dt_methods_on_dataset-1}
{\scriptsize
\begin{tabular}{llllll}
  \hline
DT Method      & Classifier           & AUC   & Precision & Recall & F-Measure \\
  \hline
RAW            & \multirow{7}{*}{DT}  & 0.524 & 0.52      & 0.526  & 0.523     \\
LOG            &                      & 0.519 & 0.515     & 0.518  & 0.516     \\
BOX-COX        &                      & 0.521 & 0.517     & 0.522  & 0.519     \\
Z-SCORE        &                      & 0.524 & 0.52      & 0.521  & 0.52      \\
RANK           &                      & 0.522 & 0.517     & 0.521  & 0.519     \\
Discritization &                      & 0.526 & 0.522     & 0.522  & 0.522     \\
WOE            &                      & 0.679 & 0.68      & 0.667  & 0.673     \\
  \hline
RAW            & \multirow{7}{*}{FNN} & 0.502 & 0.497     & 0.987  & 0.661     \\

LOG            &                      & 0.57  & 0.55      & 0.709  & 0.619     \\
BOX-COX        &                      & 0.563 & 0.549     & 0.654  & 0.597     \\
Z-SCORE        &                      & 0.708 & 0.917     & 0.917  & 0.917     \\
RANK           &                      & 0.571 & 0.556     & 0.661  & 0.604     \\
Discritization &                      & 0.566 & 0.544     & 0.743  & 0.628     \\
WOE            &                      & 0.802 & 0.799     & 0.802  & 0.8       \\
  \hline
RAW            & \multirow{7}{*}{GB}  & 0.595 & 0.584     & 0.632  & 0.607     \\

LOG            &                      & 0.591 & 0.58      & 0.631  & 0.604     \\
BOX-COX        &                      & 0.593 & 0.582     & 0.632  & 0.606     \\
Z-SCORE        &                      & 0.594 & 0.583     & 0.631  & 0.606     \\
RANK           &                      & 0.591 & 0.58      & 0.631  & 0.604     \\
Discritization &                      & 0.584 & 0.569     & 0.652  & 0.608     \\
WOE            &                      & 0.792 & 0.8       & 0.773  & 0.786     \\
  \hline
RAW            & \multirow{7}{*}{KNN} & 0.539 & 0.536     & 0.516  & 0.526     \\
 
LOG            &                      & 0.541 & 0.539     & 0.514  & 0.526     \\
BOX-COX        &                      & 0.553 & 0.55      & 0.537  & 0.543     \\
Z-SCORE        &                      & 0.541 & 0.539     & 0.506  & 0.522     \\
RANK           &                      & 0.541 & 0.539     & 0.514  & 0.526     \\
Discritization &                      & 0.544 & 0.542     & 0.519  & 0.53      \\
WOE            &                      & 0.729 & 0.856     & 0.549  & 0.669     \\
  \hline
RAW            & \multirow{7}{*}{LR}  & 0.591 & 0.588     & 0.585  & 0.586     \\
LOG            &                      & 0.577 & 0.576     & 0.558  & 0.567     \\
BOX-COX        &                      & 0.582 & 0.578     & 0.578  & 0.578     \\
Z-SCORE        &                      & 0.579 & 0.576     & 0.576  & 0.576     \\
RANK           &                      & 0.577 & 0.576     & 0.558  & 0.567     \\
Discritization &                      & 0.572 & 0.569     & 0.565  & 0.567     \\
WOE            &                      & 0.796 & 0.805     & 0.776  & 0.79      \\
  \hline
RAW            & \multirow{7}{*}{NB}  & 0.516 & 0.505     & 0.86   & 0.636     \\
LOG            &                      & 0.548 & 0.544     & 0.541  & 0.542     \\
BOX-COX        &                      & 0.53  & 0.518     & 0.709  & 0.599     \\
Z-SCORE        &                      & 0.522 & 0.509     & 0.839  & 0.634     \\
RANK           &                      & 0.548 & 0.544     & 0.541  & 0.542     \\
Discritization &                      & 0.665 & 0.52      & 1      & 0.684     \\
WOE            &                      & 0.705 & 0.797     & 0.547  & 0.649     \\
  \hline
RAW            & \multirow{7}{*}{RF}  & 0.579 & 0.572     & 0.598  & 0.585     \\
LOG            &                      & 0.588 & 0.582     & 0.596  & 0.589     \\
BOX-COX        &                      & 0.588 & 0.582     & 0.599  & 0.59      \\
Z-SCORE        &                      & 0.588 & 0.583     & 0.591  & 0.587     \\
RANK           &                      & 0.587 & 0.581     & 0.596  & 0.588     \\
Discritization &                      & 0.582 & 0.573     & 0.608  & 0.59      \\
WOE            &                      & 0.789 & 0.783     & 0.796  & 0.789     \\
  \hline
RAW            & \multirow{7}{*}{RNN} & 0.529 & 0.523     & 0.549  & 0.536     \\
LOG            &                      & 0.535 & 0.525     & 0.632  & 0.574     \\
BOX-COX        &                      & 0.554 & 0.546     & 0.594  & 0.569     \\
Z-SCORE        &                      & 0.552 & 0.539     & 0.657  & 0.592     \\
RANK           &                      & 0.537 & 0.525     & 0.671  & 0.589     \\
Discritization &                      & 0.549 & 0.542     & 0.574  & 0.558     \\
WOE            &                      & 0.714 & 0.712     & 0.709  & 0.71     \\
  \hline

\end{tabular}
}
\end{center}
\end{table}

\begin{table}[]
\begin{center}
\caption{RAW and  DT method based CCP models on dataset-2}
\label{table:dt_methods_on_dataset-2}
{\scriptsize
\begin{tabular}{llllll}
  \hline
DT Method      & Classifier            & AUC   & Precision & Recall & F-Measure \\
 \hline
RAW            &                       & 0.826 & 0.679     & 0.707  & 0.693     \\
LOG            &                       & 0.823 & 0.683     & 0.699  & 0.691     \\
BOX-COX        &                       & 0.83  & 0.698     & 0.711  & 0.704     \\
Z-SCORE        &                       & 0.834 & 0.692     & 0.72   & 0.706     \\
RANK           &                       & 0.798 & 0.636     & 0.659  & 0.647     \\
Discritization &                       & 0.815 & 0.666     & 0.686  & 0.676     \\
WOE            & \multirow{-7}{*}{DT}  & 0.777 & 0.652     & 0.607  & 0.629     \\
 \hline
RAW            &                       & 0.5   & 0.142     & 0.1    & 0.117     \\
LOG            &                       & 0.515 & 0.5       & 0.037  & 0.069     \\
BOX-COX        &                       & 0.501 & 0.5       & 0.001  & 0.002     \\
Z-SCORE        &                       & 0.653 & 0.588     & 0.345  & 0.435     \\
RANK           &                       & 0.615 & 0.537     & 0.267  & 0.357     \\
Discritization &                       & 0.578 & 0.575     & 0.178  & 0.272     \\
WOE            & \multirow{-7}{*}{FNN} & 0.819 & 0.816     & 0.663  & 0.732     \\
 \hline
RAW            &                       & 0.836 & 0.91      & 0.683  & 0.78      \\
LOG            &                       & 0.835 & 0.899     & 0.683  & 0.776     \\
BOX-COX        &                       & 0.835 & 0.901     & 0.682  & 0.776     \\
Z-SCORE        &                       & 0.833 & 0.894     & 0.679  & 0.772     \\
RANK           &                       & 0.814 & 0.856     & 0.646  & 0.736     \\
Discritization &                       & 0.824 & 0.901     & 0.659  & 0.761     \\
WOE            & \multirow{-7}{*}{GB}  & 0.82  & 0.835     & 0.661  & 0.738     \\
 \hline
RAW            &                       & 0.616 & 0.625     & 0.257  & 0.364     \\
LOG            &                       & 0.532 & 0.486     & 0.076  & 0.131     \\
BOX-COX        &                       & 0.587 & 0.825     & 0.18   & 0.296     \\
Z-SCORE        &                       & 0.555 & 0.703     & 0.117  & 0.201     \\
RANK           &                       & 0.606 & 0.75      & 0.225  & 0.346     \\
Discritization &                       & 0.566 & 0.739     & 0.14   & 0.235     \\
WOE            & \multirow{-7}{*}{KNN} & 0.773 & 0.867     & 0.56   & 0.68      \\
 \hline
RAW            &                       & 0.566 & 0.576     & 0.15   & 0.238     \\
LOG            &                       & 0.56  & 0.594     & 0.134  & 0.219     \\
BOX-COX        &                       & 0.571 & 0.595     & 0.16   & 0.252     \\
Z-SCORE        &                       & 0.572 & 0.557     & 0.165  & 0.255     \\
RANK           &                       & 0.567 & 0.598     & 0.151  & 0.241     \\
Discritization &                       & 0.568 & 0.552     & 0.157  & 0.244     \\
WOE            & \multirow{-7}{*}{LR}  & 0.828 & 0.84      & 0.678  & 0.75      \\
 \hline
RAW            &                       & 0.67  & 0.492     & 0.41   & 0.447     \\
LOG            &                       & 0.543 & 0.624     & 0.096  & 0.166     \\
BOX-COX        &                       & 0.579 & 0.577     & 0.18   & 0.274     \\
Z-SCORE        &                       & 0.6   & 0.628     & 0.222  & 0.328     \\
RANK           &                       & 0.592 & 0.673     & 0.201  & 0.31      \\
Discritization &                       & 0.62  & 0.656     & 0.262  & 0.374     \\
WOE            & \multirow{-7}{*}{NB}  & 0.795 & 0.762     & 0.621  & 0.684     \\
 \hline
RAW            &                       & 0.507 & 0.833     & 0.014  & 0.028     \\
LOG            &                       & 0.829 & 0.927     & 0.666  & 0.775     \\
BOX-COX        &                       & 0.83  & 0.931     & 0.669  & 0.779     \\
Z-SCORE        &                       & 0.832 & 0.917     & 0.673  & 0.776     \\
RANK           &                       & 0.803 & 0.873     & 0.621  & 0.726     \\
Discritization &                       & 0.821 & 0.904     & 0.653  & 0.758     \\
WOE            & \multirow{-7}{*}{RF}  & 0.796 & 0.862     & 0.608  & 0.713     \\
 \hline
RAW            &                       & 0.509 & 0.253     & 0.034  & 0.06      \\
LOG            &                       & 0.501 & 0.5       & 0.001  & 0.002     \\
BOX-COX        &                       & 0.501 & 0.5       & 0.001  & 0.002     \\
Z-SCORE        &                       & 0.501 & 0.5       & 0.001  & 0.002     \\
RANK           &                       & 0.502 & 0.75      & 0.004  & 0.008     \\
Discritization &                       & 0.501 & 0.5       & 0.001  & 0.002     \\
WOE            & \multirow{-7}{*}{RNN} & 0.668 & 0.859     & 0.345  & 
 0.492    \\
  \hline
 
\end{tabular}
}
\end{center}
\end{table}

\begin{table}[]
\begin{center}
\caption{RAW and  DT method based CCP models on dataset-3}
\label{table:dt_methods_on_dataset-3}
{\scriptsize
\begin{tabular}{llllll}
   \hline
DT   method    & Classifier            & AUC   & Precision & Recall & F-Measure \\
  \hline
RAW            &                       & 0.84  & 0.729     & 0.725  & 0.727     \\
LOG            &                       & 0.843 & 0.737     & 0.731  & 0.734     \\
BOX-COX        &                       & 0.84  & 0.717     & 0.729  & 0.723     \\
Z-SCORE        &                       & 0.842 & 0.726     & 0.731  & 0.728     \\
RANK           &                       & 0.819 & 0.692     & 0.689  & 0.69      \\
Discritization &                       & 0.845 & 0.704     & 0.743  & 0.723     \\
WOE            & \multirow{-7}{*}{DT}  & 0.803 & 0.695     & 0.654  & 0.674     \\
  \hline
RAW            &                       & 0.506 & 0.158     & 0.112  & 0.131     \\
LOG            &                       & 0.529 & 0.582     & 0.066  & 0.119     \\
BOX-COX        &                       & 0.501 & 0.333     & 0.004  & 0.008     \\
Z-SCORE        &                       & 0.594 & 0.522     & 0.224  & 0.313     \\
RANK           &                       & 0.575 & 0.518     & 0.178  & 0.265     \\
Discritization &                       & 0.568 & 0.619     & 0.151  & 0.243     \\
WOE            & \multirow{-7}{*}{FNN} & 0.841 & 0.811     & 0.71   & 0.757     \\
  \hline
RAW            &                       & 0.86  & 0.899     & 0.735  & 0.809     \\
LOG            &                       & 0.864 & 0.915     & 0.739  & 0.818     \\
BOX-COX        &                       & 0.858 & 0.9       & 0.729  & 0.806     \\
Z-SCORE        &                       & 0.857 & 0.898     & 0.729  & 0.805     \\
RANK           &                       & 0.844 & 0.857     & 0.708  & 0.775     \\
Discritization &                       & 0.851 & 0.889     & 0.716  & 0.793     \\
WOE            & \multirow{-7}{*}{GB}  & 0.826 & 0.84      & 0.673  & 0.747     \\
  \hline
RAW            &                       & 0.621 & 0.676     & 0.263  & 0.379     \\
LOG            &                       & 0.548 & 0.619     & 0.108  & 0.184     \\
BOX-COX        &                       & 0.587 & 0.977     & 0.174  & 0.295     \\
Z-SCORE        &                       & 0.591 & 0.835     & 0.188  & 0.307     \\
RANK           &                       & 0.575 & 0.815     & 0.155  & 0.26      \\
Discritization &                       & 0.62  & 0.801     & 0.251  & 0.382     \\
WOE            & \multirow{-7}{*}{KNN} & 0.776 & 0.81      & 0.576  & 0.673     \\
  \hline
RAW            &                       & 0.58  & 0.567     & 0.184  & 0.278     \\
LOG            &                       & 0.564 & 0.579     & 0.145  & 0.232     \\
BOX-COX        &                       & 0.58  & 0.535     & 0.188  & 0.278     \\
Z-SCORE        &                       & 0.592 & 0.573     & 0.211  & 0.308     \\
RANK           &                       & 0.568 & 0.577     & 0.155  & 0.244     \\
Discritization &                       & 0.57  & 0.549     & 0.161  & 0.249     \\
WOE            & \multirow{-7}{*}{LR}  & 0.839 & 0.819     & 0.704  & 0.757     \\
  \hline
RAW            &                       & 0.728 & 0.533     & 0.536  & 0.534     \\
LOG            &                       & 0.538 & 0.621     & 0.085  & 0.15      \\
BOX-COX        &                       & 0.603 & 0.806     & 0.215  & 0.339     \\
Z-SCORE        &                       & 0.655 & 0.679     & 0.337  & 0.45      \\
RANK           &                       & 0.614 & 0.692     & 0.246  & 0.363     \\
Discritization &                       & 0.69  & 0.65      & 0.418  & 0.509     \\
WOE            & \multirow{-7}{*}{NB}  & 0.797 & 0.761     & 0.627  & 0.688     \\
  \hline
RAW            &                       & 0.51  & 0.909     & 0.021  & 0.041     \\
LOG            &                       & 0.858 & 0.948     & 0.723  & 0.82      \\
BOX-COX        &                       & 0.856 & 0.943     & 0.718  & 0.815     \\
Z-SCORE        &                       & 0.851 & 0.945     & 0.708  & 0.81      \\
RANK           &                       & 0.827 & 0.897     & 0.667  & 0.765     \\
Discritization &                       & 0.836 & 0.929     & 0.681  & 0.786     \\
WOE            & \multirow{-7}{*}{RF}  & 0.797 & 0.87      & 0.609  & 0.716     \\
  \hline
RAW            &                       & 0.501 & 0.5       & 0.002  & 0.004     \\
LOG            &                       & 0.501 & 0.5       & 0.002  & 0.004     \\
BOX-COX        &                       & 0.501 & 0.5       & 0.002  & 0.004     \\
Z-SCORE        &                       & 0.501 & 0.5       & 0.002  & 0.004     \\
RANK           &                       & 0.501 & 0.5       & 0.002  & 0.004     \\
Discritization &                       & 0.501 & 0.5       & 0.002  & 0.004     \\
WOE            & \multirow{-7}{*}{RNN} & 0.672 & 0.933     & 0.348  & 0.507    \\
  \hline
\end{tabular}
}
\end{center}
\end{table}

\end{document}